\documentclass[conference]{IEEEtran}
\IEEEoverridecommandlockouts

\newif\ifarxiv
\arxivtrue

\newif\iffinal
\finaltrue

\newif\ifieee
\ieeefalse

\usepackage{cite}
\usepackage{amsmath,amssymb,amsfonts}
\usepackage{algorithmic}
\usepackage{graphicx}
\usepackage{textcomp}

\usepackage[caption=false,farskip=0pt]{subfig}
\usepackage{booktabs}
\usepackage{multirow}
\usepackage[hyphens]{url}
\usepackage{balance}

\ifieee
\else
\usepackage[colorlinks=true,allcolors=black]{hyperref} %
\fi

\usepackage[acronym]{glossaries}
\usepackage{xspace}
\usepackage[utf8]{inputenc}
\usepackage[T1]{fontenc}
\usepackage[dvipsnames,table]{xcolor}
\usepackage{diagbox}
\usepackage[capitalise]{cleveref} %
\usepackage{comment}

\usepackage{soul}

\newcounter{fncounter}
\newcommand\customfootnote[1]{\stepcounter{fncounter}\footnote{\hspace{0.2mm}#1}}

\newcommand*{\RL}[2][]{\textcolor{Rhodamine}{[\textbf{\ifthenelse{\equal{#1}{}}{RL}{RL(#1)}}: #2]}}
\newcommand\RLI[1]{} %
\newcommand*{\DM}[2][]{\textcolor{blue}{[\textbf{\ifthenelse{\equal{#1}{}}{DM}{DM(#1)}}: #2]}}

\newacronymstyle{long-short-br}
{%
  \GlsUseAcrEntryDispStyle{long-short}%
}%
{%
  \GlsUseAcrStyleDefs{long-short}%
}
\setacronymstyle{long-short-br}


\usepackage{transparent}
\usepackage{tikz}
\ifarxiv
\newcommand\copyrighttext{%
  \scriptsize Accepted at IJCNN 2025. The final published version is available on IEEE Xplore (DOI: \href{https://doi.org/10.1109/IJCNN64981.2025.11227421}{\textcolor{blue}{10.1109/IJCNN64981.2025.11227421}}).}
\newcommand\copyrightnotice{%
\begin{tikzpicture}[remember picture,overlay]
\node[anchor=south,yshift=30pt,xshift=0pt] at (current page.south) {\fbox{\transparent{0.85}\parbox{\dimexpr0.72\textwidth-\fboxsep-\fboxrule\relax}{\copyrighttext}}};
\end{tikzpicture}%
}

\else
\fi

\newcommand{\papertitle}{Improving Small Drone Detection Through Multi-Scale Processing and Data~Augmentation}

\begin{document}

\newacronym{iou}{IoU}{Intersection over Union}
\newacronym{nms}{NMS}{Non-Maximum Suppression}
\newacronym{sgd}{SGD}{Stochastic Gradient Descent}

\newacronym{map}{mAP}{mean Average Precision}

\newacronym{dds}{DDS}{Drone Data Set}
\newacronym{usc}{USC-GRAD-STDdb}{Small Target Detection database}
\newacronym{ijcnn}{IJCNN}{International Joint Conference on Neural Networks}
\newacronym{wosdetc}{WOSDETC}{International Workshop on Small-Drone Surveillance, Detection and Counteraction Techniques}
\newacronym{uav}{UAV}{Unmanned Aerial Vehicle}

\newcommand{\dds}{\gls*{dds}\xspace}
\newcommand{\dut}{DUT Anti-UAV\xspace}
\newcommand{\stddb}{\usc\xspace}
\newcommand{\usc}{\gls*{usc}\xspace}

\newcommand{\model}{YOLO11m\xspace}
\newcommand{\mobilenet}{MobileNetV3\xspace}
\newcommand{\stdnet}{STDnet\xspace}

\newcommand{\challenge}{8th WOSDETC Drone-vs-Bird Detection Grand Challenge\xspace}
\iffinal
\newcommand{\supplementary}{\url{https://raysonlaroca.github.io/supp/drone-vs-bird/}}
\newcommand{\supplementaryEndParagraph}{\supplementary}
\else
\newcommand{\supplementary}{[\textit{hidden for review}]}
\newcommand{\supplementaryEndParagraph}{[\textit{hidden for~review}]}
\fi

\iffinal
\title{\papertitle
}
\else
\title{\papertitle
\thanks{The funding agencies are hidden for review.}
}
\fi

\iffinal
\author{Rayson Laroca\IEEEauthorrefmark{1}$^,$\IEEEauthorrefmark{2}, Marcelo dos Santos\IEEEauthorrefmark{2}, and David Menotti\IEEEauthorrefmark{2}\\[5pt]
\IEEEauthorrefmark{1}\hspace{0.15mm}Pontifical Catholic University of Paran\'a, Curitiba, Brazil\\
\IEEEauthorrefmark{2}\hspace{0.15mm}Federal University of Paran\'a, Curitiba, Brazil\\[1ex]
\IEEEauthorrefmark{1}{\hspace{0.15mm}\tt\small \texttt{rayson@ppgia.pucpr.br}} \quad \IEEEauthorrefmark{2}{\hspace{-0.2mm}\tt\small\{msantos, menotti\}@inf.ufpr.br}}
\else
\author{\IEEEauthorblockN{Anonymous Authors}}
\fi

\maketitle

\ifarxiv
    \copyrightnotice
\else
\fi

\glsresetall
\ifarxiv
  \vspace{-3.575mm}
\else
\fi
\begin{abstract}
Detecting small drones, often indistinguishable from birds, is crucial for modern surveillance.
This work introduces a drone detection methodology built upon the medium-sized YOLOv11 object detection model.
To enhance its performance on small targets, we implemented a multi-scale approach in which the input image is processed both as a whole and in segmented parts, with subsequent prediction aggregation.
We also utilized a copy-paste data augmentation technique to enrich the training dataset with diverse drone and bird examples.
Finally, we implemented a post-processing technique that leverages frame-to-frame consistency to mitigate missed detections.
The proposed approach attained first place in the \challenge, held at the 2025~\gls*{ijcnn}, showcasing its capability to detect drones in complex environments~effectively.

\end{abstract}

\section{Introduction}
\label{sec:introduction}

\glsresetall

\glspl*{uav}, commonly known as drones, have experienced a surge in popularity across diverse civil sectors~\cite{liu2020unmanned,mohsan2023unmanned}.
Their autonomy, flexibility, and affordability have driven their widespread adoption in applications such as search and rescue, package delivery, and remote sensing~\cite{zhao2023workshop}.
However, this rapid expansion presents significant security and privacy challenges~\cite{yahuza2021internet,chamola2021comprehensive}.

The versatility of drones also makes them susceptible to malicious uses, such as smuggling contraband, conducting intrusive surveillance, and executing physical attacks~\cite{kunertova2022ukraine,krame2023narco}.
Additionally, unauthorized drone operations can violate aviation safety regulations, posing direct threats to civilian aircraft and passengers while causing disruptions at airports, including flight delays~\cite{wendt2020estimating,park2021survey}.
As a result, there is an escalating and urgent demand for advanced drone detection systems to address these growing security and privacy concerns~\cite{gao2024novel,munir2024investigation}.

\begin{figure}[!t]
    \vspace{0.8mm}
    \centering
    \includegraphics[width=0.975\linewidth]{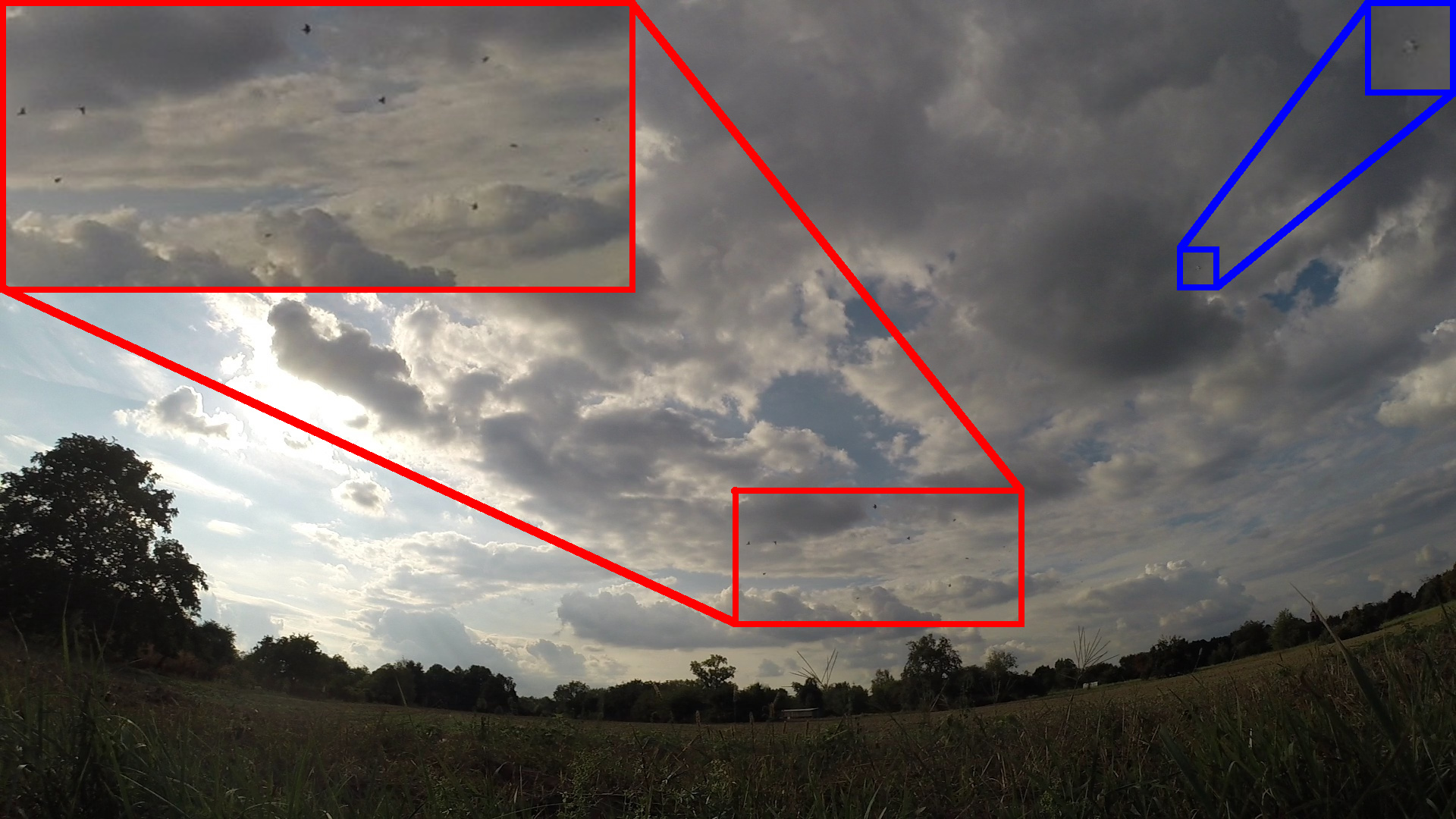}

    \vspace{-2mm}
    
    \caption{The similarity in size and appearance often makes it challenging to differentiate a drone~(blue box) from birds~(red box), especially at a~distance.}
    \label{fig:sample-drone-vs-bird}
\end{figure}

A major challenge in drone detection is accurately distinguishing drones from birds, given their similarities in size and appearance (see \cref{fig:sample-drone-vs-bird}).
Moreover, practical systems must be capable of identifying drones at long distances to allow for timely responses.
This requires detecting very small objects and differentiating them from structured backgrounds and other complex visual elements.
To spur innovation in this area, the \gls*{wosdetc} consortium has launched the \challenge at the 2025 \gls*{ijcnn}~\cite{challenge}.
The challenge seeks to attract research focused on developing novel signal-processing solutions for the problem of distinguishing birds from drones at long distances.
To support participants, the consortium provides a video dataset, which is inherently difficult to acquire given the specific conditions and permissions required for drone flights. 
The dataset is progressively expanded with each challenge installment~\cite{coluccia2021drone,coluccia2024drone} and subsequently released to the research~community.

Considering the preceding discussion, this work presents a drone detection methodology that leverages the state-of-the-art YOLOv11 object detection model~\cite{yolov11}.
To enhance the detection of small-scale drones, we implemented a multi-scale approach that processes the input image both as a whole and in segmented components (simulating a zoom effect).
Additionally, we employed extensive data augmentation, including a copy-paste technique, to increase the representation of drones and birds in the training images.
A post-processing stage, utilizing adjacent frame predictions, was also employed to reduce the number of missed detections.
Notably, our approach was awarded first place in the \challenge~\cite{challenge}.

The rest of this paper is organized as follows: 
\cref{sec:related_work} briefly reviews related work.
\cref{sec:experiments} describes the experiments conducted, including the datasets explored, the proposed method, and the experimental results.
Finally, \cref{sec:conclusions} concludes the paper and suggests directions for future~research.

\section{Related Work}
\label{sec:related_work}

To contextualize our work within the emerging yet rapidly growing field of drone detection, this section focuses on reviewing the works that introduced the datasets employed in our experiments, as well as the top-performing methods from the previous installment of the Drone-vs-Bird challenge~\cite{coluccia2024drone}.

Bosquet et al.~\cite{bosquet2018stdnet} highlighted the lack of networks specifically designed for detecting small targets, defining small objects as those smaller than $16\times16$ pixels.
They then introduced \stdnet, a region-proposal-based network that incorporates a visual attention mechanism to select the most promising regions in the feature map while disregarding less relevant areas.
Evaluated on their proposed \acrshort*{usc} dataset, \stdnet demonstrated superior performance compared to Faster R-CNN~\cite{ren2017faster} and FPN~\cite{lin2017feature}.

Svanstr\"{o}m et al.~\cite{svanstrom2021real} explored sensor fusion for drone detection, an area with limited research at the time.
They developed a multi-sensor system that employed then state-of-the-art models, such as YOLOv2~\cite{redmon2017yolo9000}, and combined class outputs and confidence scores from multiple sensors, including infrared cameras, visible cameras, and audio.
Their analysis of camera data identified insects and clouds as the primary sources of false positives.
Remarkably, implementing a two-sensor confirmation protocol significantly reduced these~errors.

Zhao et al.~\cite{zhao2022vision} advanced anti-UAV research by evaluating state-of-the-art object detectors and trackers on their proposed \dut dataset, which comprises 10,000 images and 20 sequences captured in diverse scenarios.
They explored combinations of 14 detectors, including Faster-RCNN~\cite{ren2017faster} and YOLOX\cite{ge2021yolox}, along with 8 trackers, such as SiamFC~\cite{bertinetto2016fully} and TransT~\cite{chen2021transformer}.
Additionally, they proposed a fusion strategy that improves tracking robustness by integrating detection.
When tracker confidence falls below a threshold, the detector refines the bounding box, ensuring more accurate~tracking.

OBSS AI (OBSS Teknoloji, Ankara, Turkey), the winning team of the previous Drone-vs-Bird challenge~\cite{coluccia2024drone}, proposed a drone detection framework comprising an object detector, a sequence classifier, and a template matching module.
They employed YOLOv5m6~\cite{yolov5} for object detection, followed by an object tracker to generate trajectories for detected objects.
The sequence classifier then evaluated eight instances from each track to estimate the probability of a drone.
To address missed detections in complex scenes, the framework employed template matching around the last known location within a small search region, producing bounding boxes that were subsequently evaluated by the sequence classifier.
Although their approach was not formally published, it was heavily inspired by the team's earlier work~\cite{akyon2022sequence}.

Mistry et al.~\cite{mistry2023drone}, the runner-up team in the previous installment of the Drone-vs-Bird challenge~\cite{coluccia2024drone}, developed a drone detection algorithm featuring three key stages.
First, YOLOv7~\cite{wang2023yolov7} performed initial detection.
Second, false positives were reduced using a heuristic based on confidence score thresholds.
Specifically, the algorithm estimates the number of drones,~$n$, in each sequence and retains only the~$n$ bounding boxes with the highest confidence scores in each frame.
Finally, a CSRT tracker~\cite{lukezic2018discriminative} was employed to reduce missed detections in complex~environments.

Kim et al.~\cite{kim2023high}, who placed third in the previous Drone-vs-Bird challenge~\cite{coluccia2024drone}, chose the YOLOv8 model~\cite{yolov8} for drone detection.
Recognizing the challenge of detecting small drones, they enhanced YOLOv8 by incorporating the P2 layer from its backbone into the feature pyramid, thereby leveraging its detailed spatial information.
They further improved inference performance by scaling input images to 1280 pixels and employing extensive data~augmentation.

We developed our drone detection method by combining key analytical insights from prior research.
Specifically, we employ a state-of-the-art iteration of the YOLO family, given its strong track record.
Our approach includes multi-scale processing and extensive data augmentation techniques to enhance model generalization and robustness, particularly in detecting small-scale drones.
Lastly, we integrate temporal information to reduce missed detections and improve overall~performance.
\section{Experiments}
\label{sec:experiments}

This section describes the experiments conducted for this work.
\cref{sec:experiments:datasets} provides details on the datasets used for training and validation.
\cref{sec:experiments:proposed} elaborates on the design and implementation of the proposed approach.
Lastly, \cref{sec:experiments:results} presents the experimental~results.

All experiments were performed on a PC equipped with an AMD Ryzen Threadripper $1920$X CPU ($3.5$GHz), $64$~GB of RAM, and an NVIDIA Quadro RTX~$8000$ GPU ($48$~GB).

\subsection{Datasets}
\label{sec:experiments:datasets}

\begin{figure}[!htb]
    \centering

    \resizebox{0.99\linewidth}{!}{
        \includegraphics[height=9.5ex]{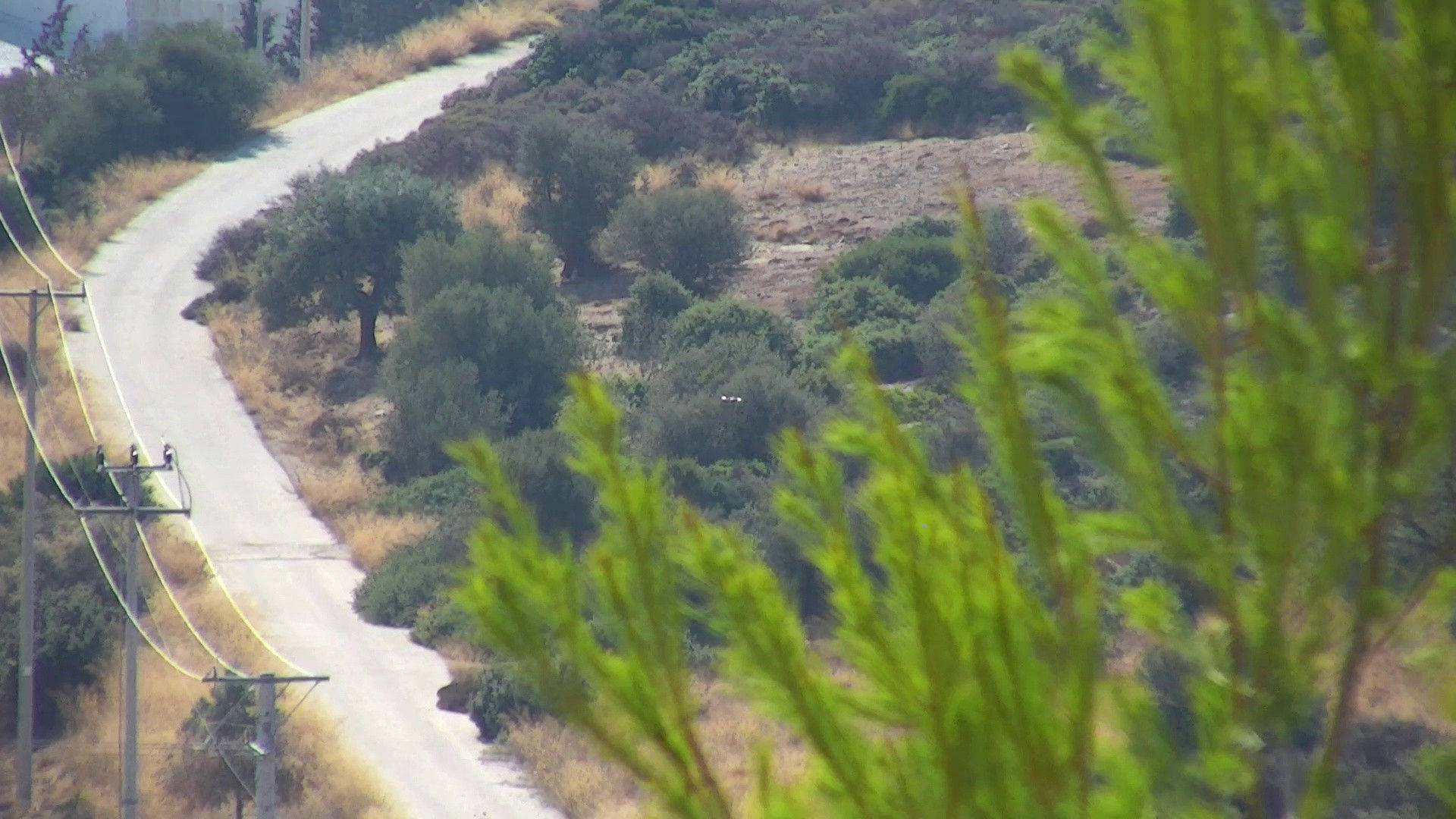}
        \includegraphics[height=9.5ex]{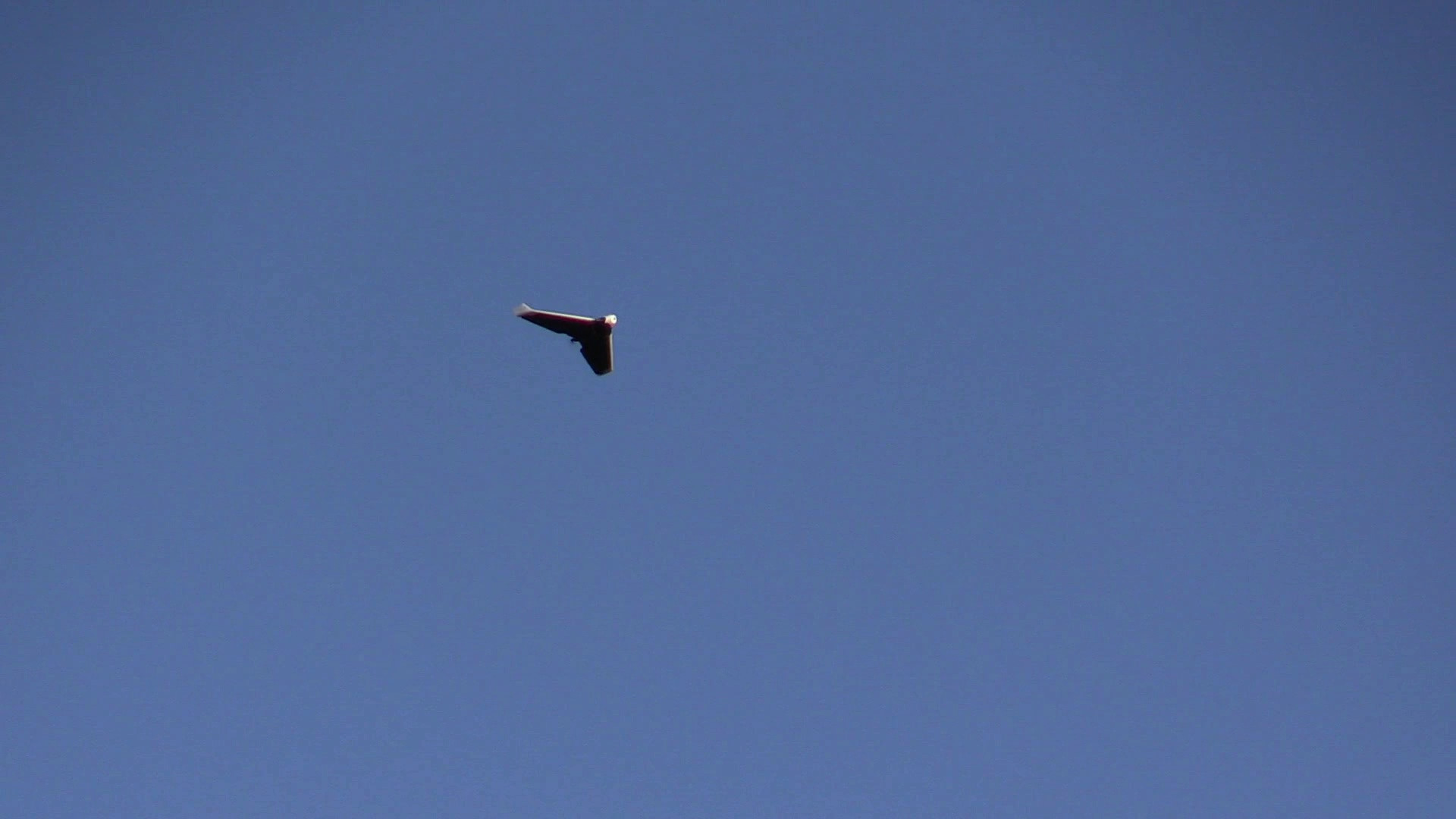}
        \includegraphics[height=9.5ex]{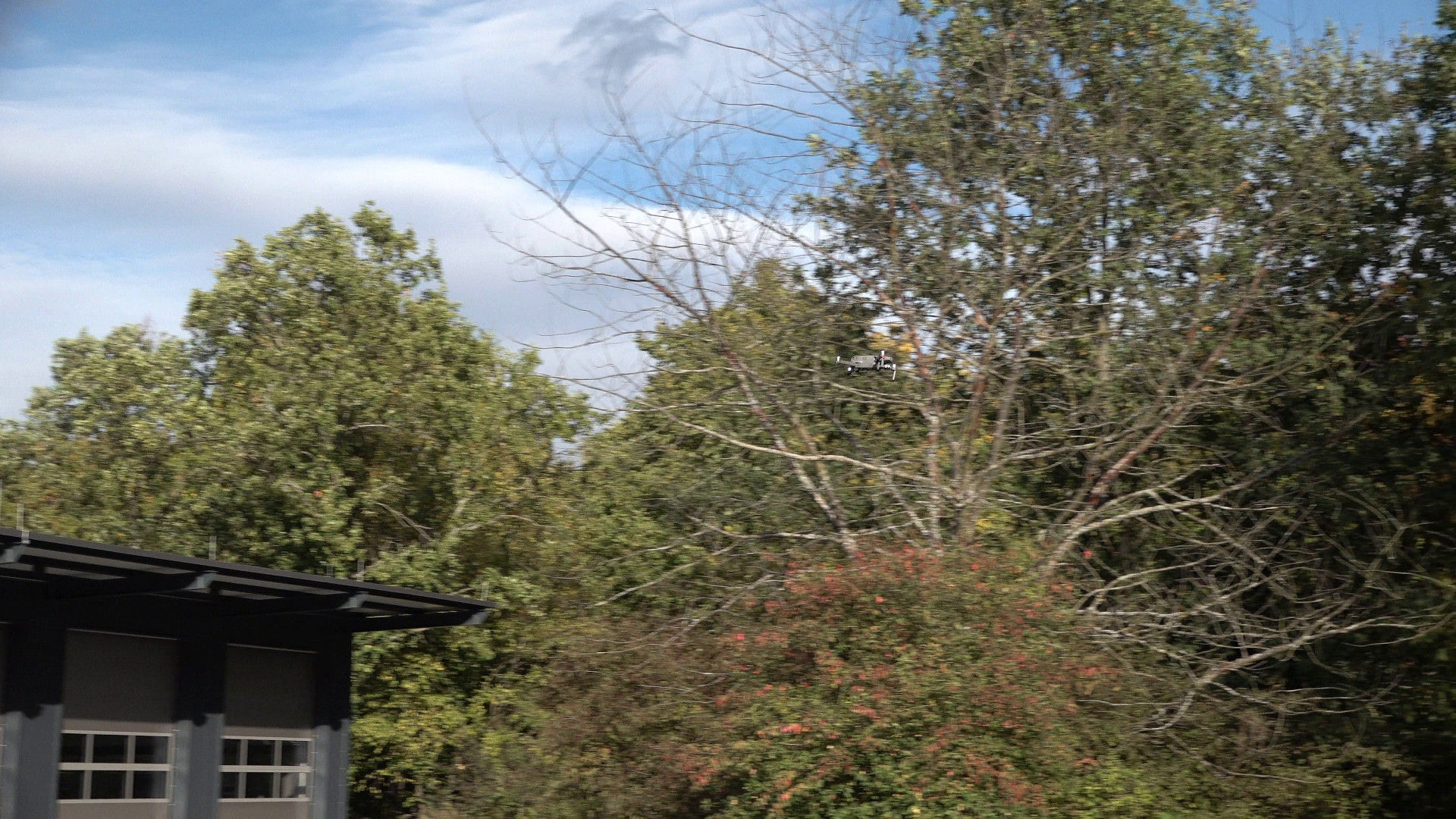}
    }

    \vspace{1mm}

    \resizebox{0.99\linewidth}{!}{
        \includegraphics[height=10.5ex]{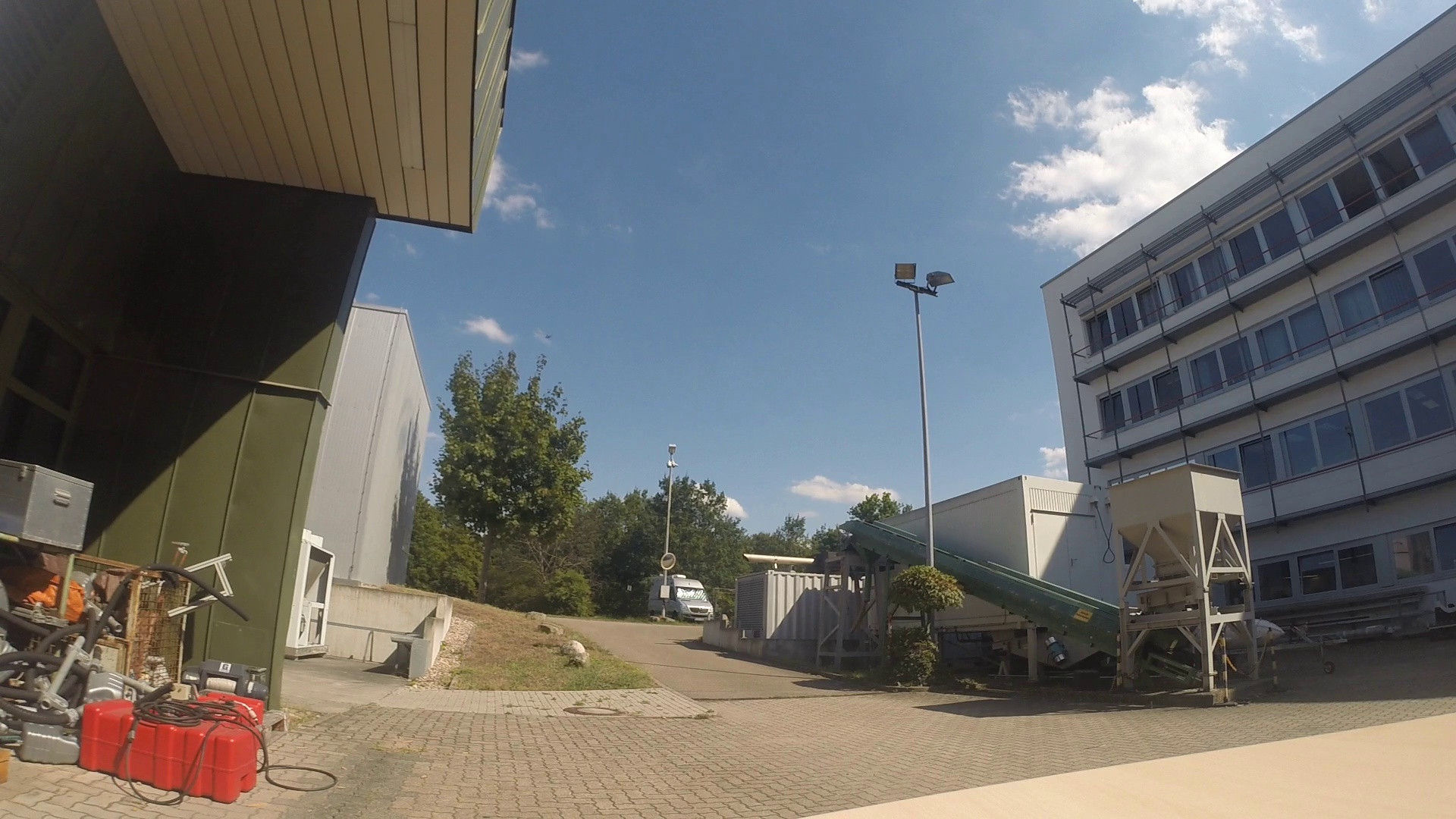}
        \includegraphics[height=10.5ex]{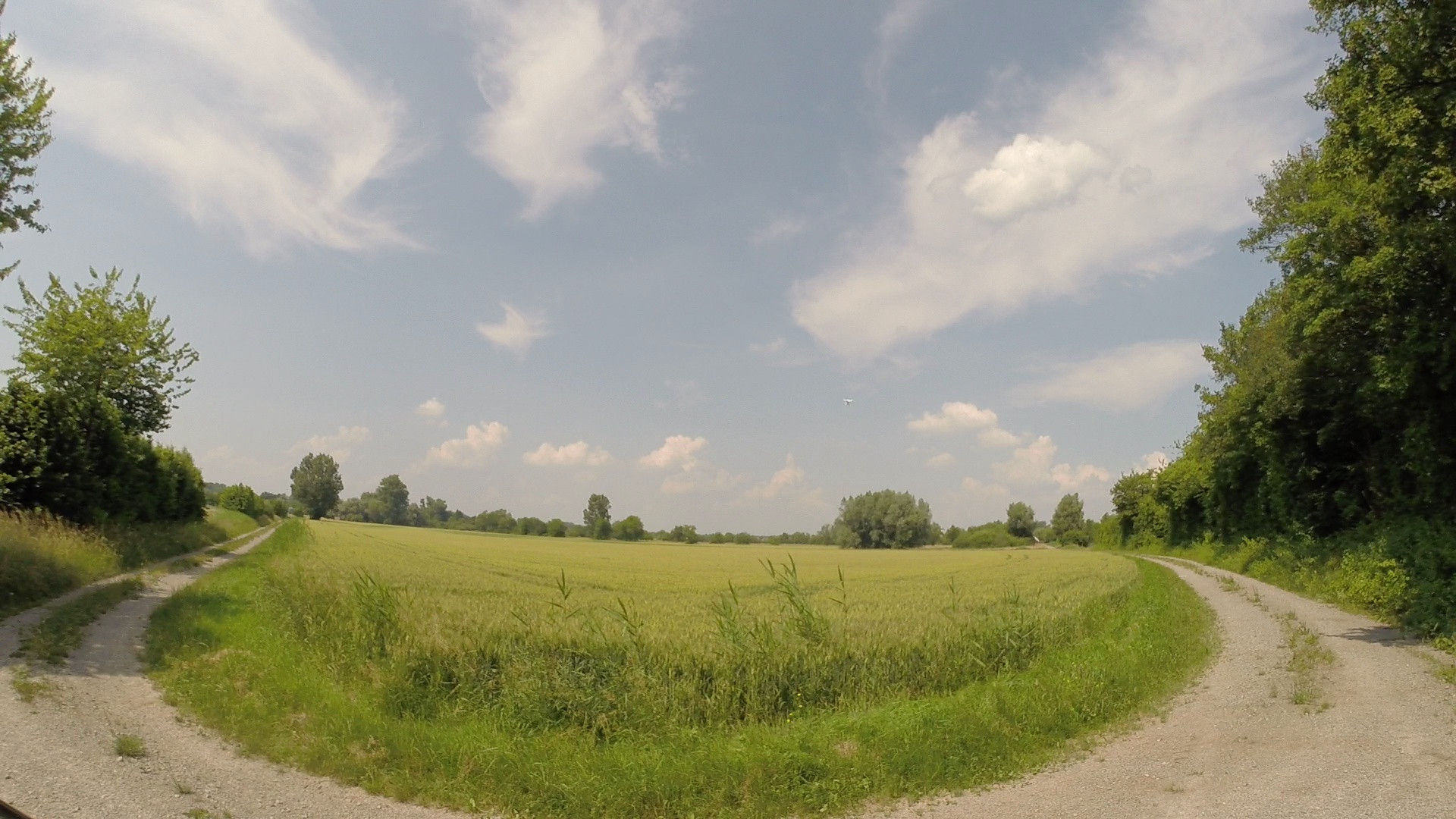}
        \includegraphics[height=10.5ex]{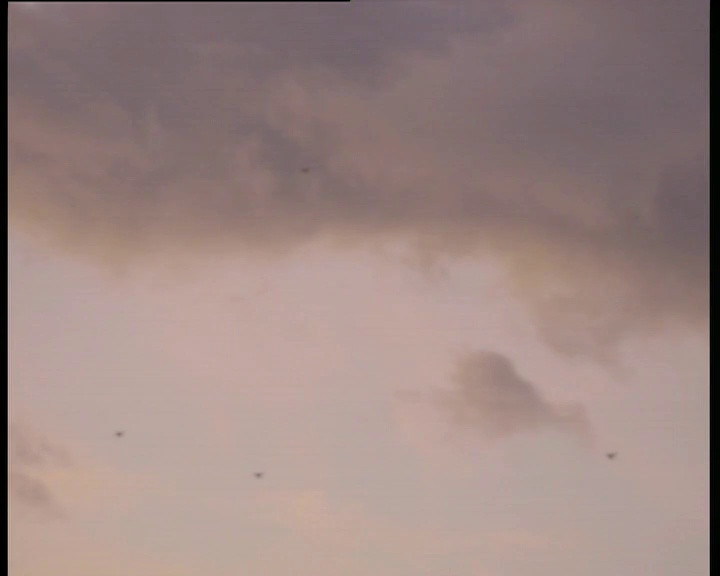}
    }

    \vspace{1mm}

    \resizebox{0.99\linewidth}{!}{
        \includegraphics[height=9.5ex]{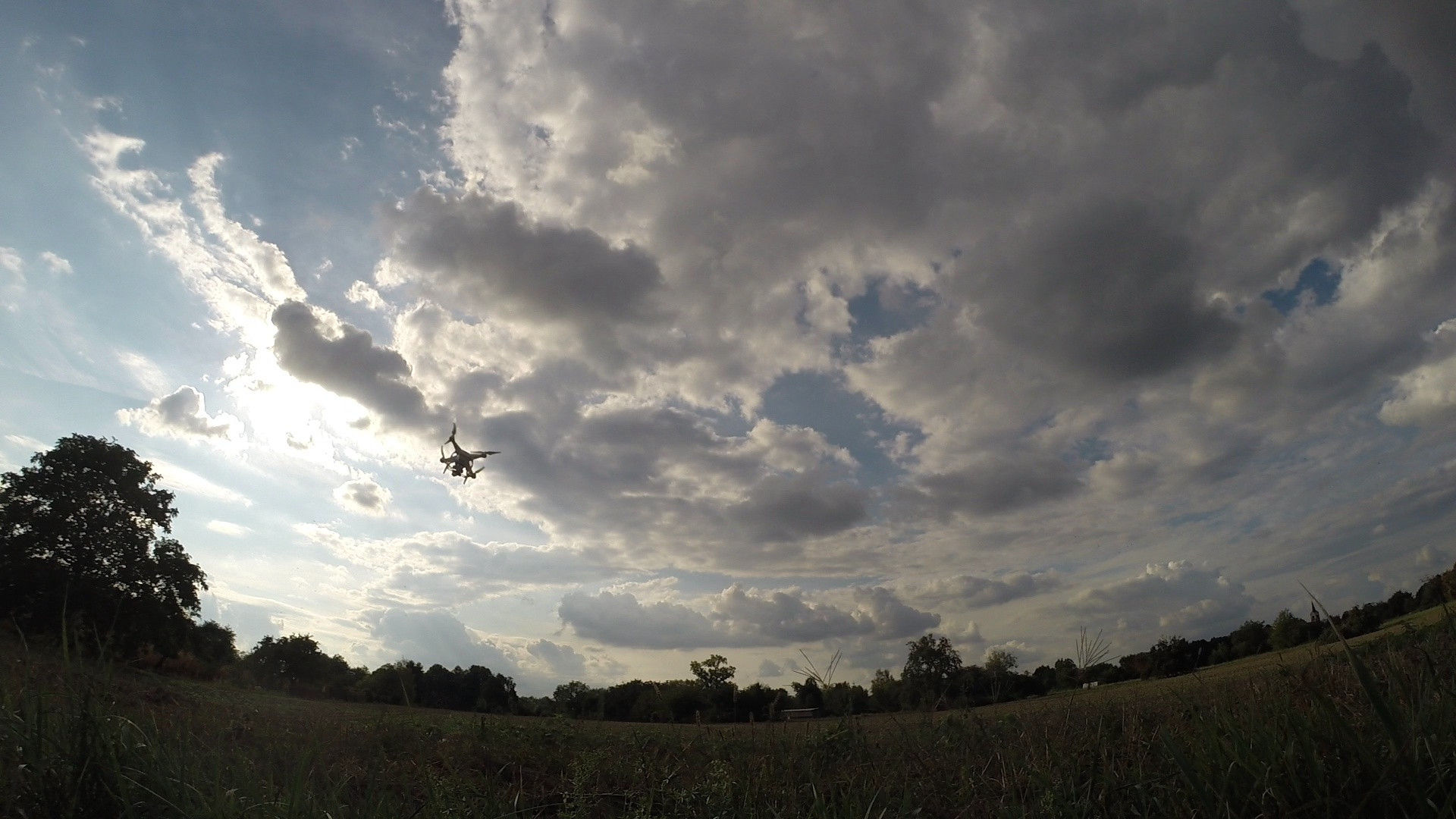}
        \includegraphics[height=9.5ex]{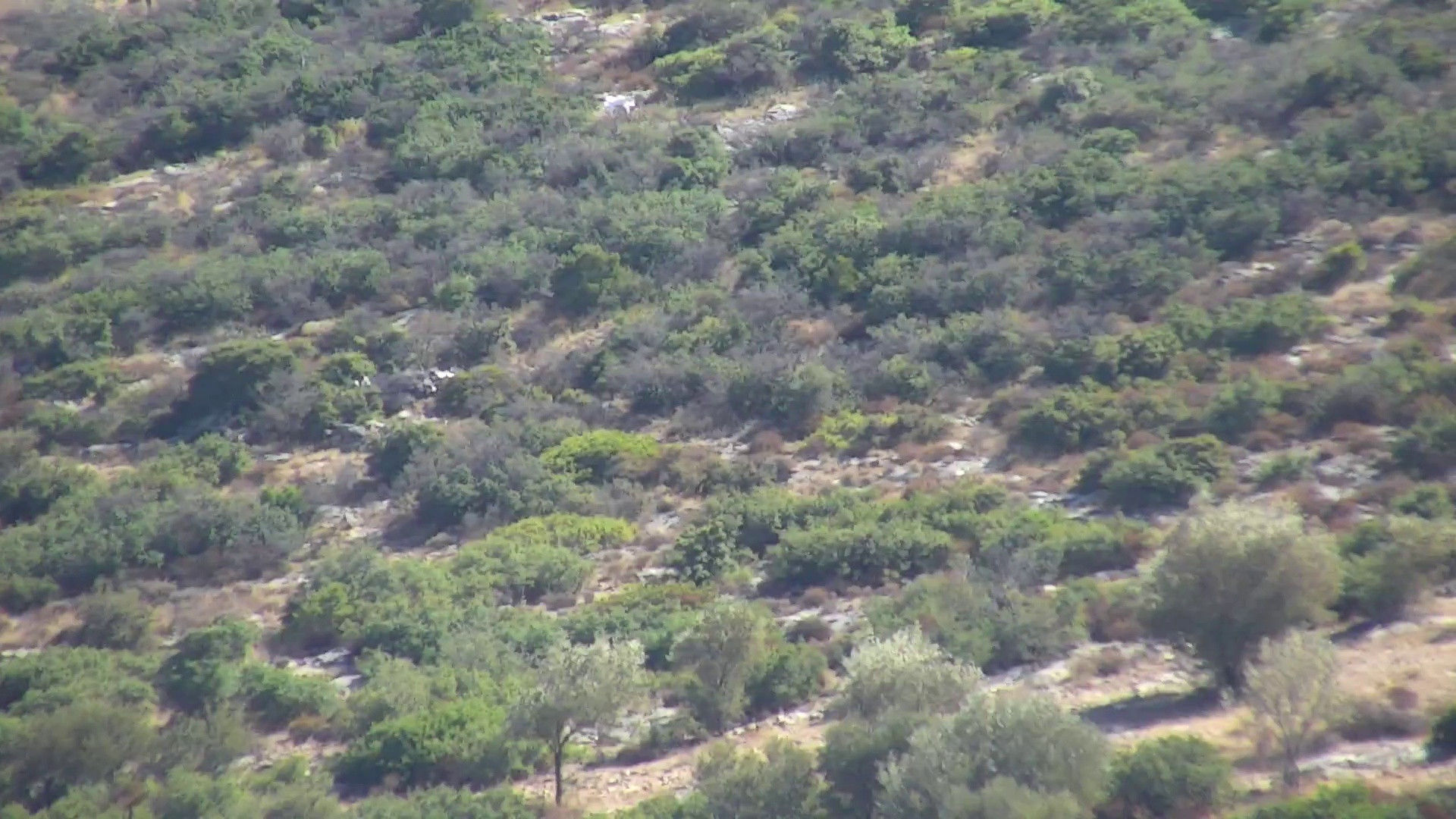}
        \includegraphics[height=9.5ex]{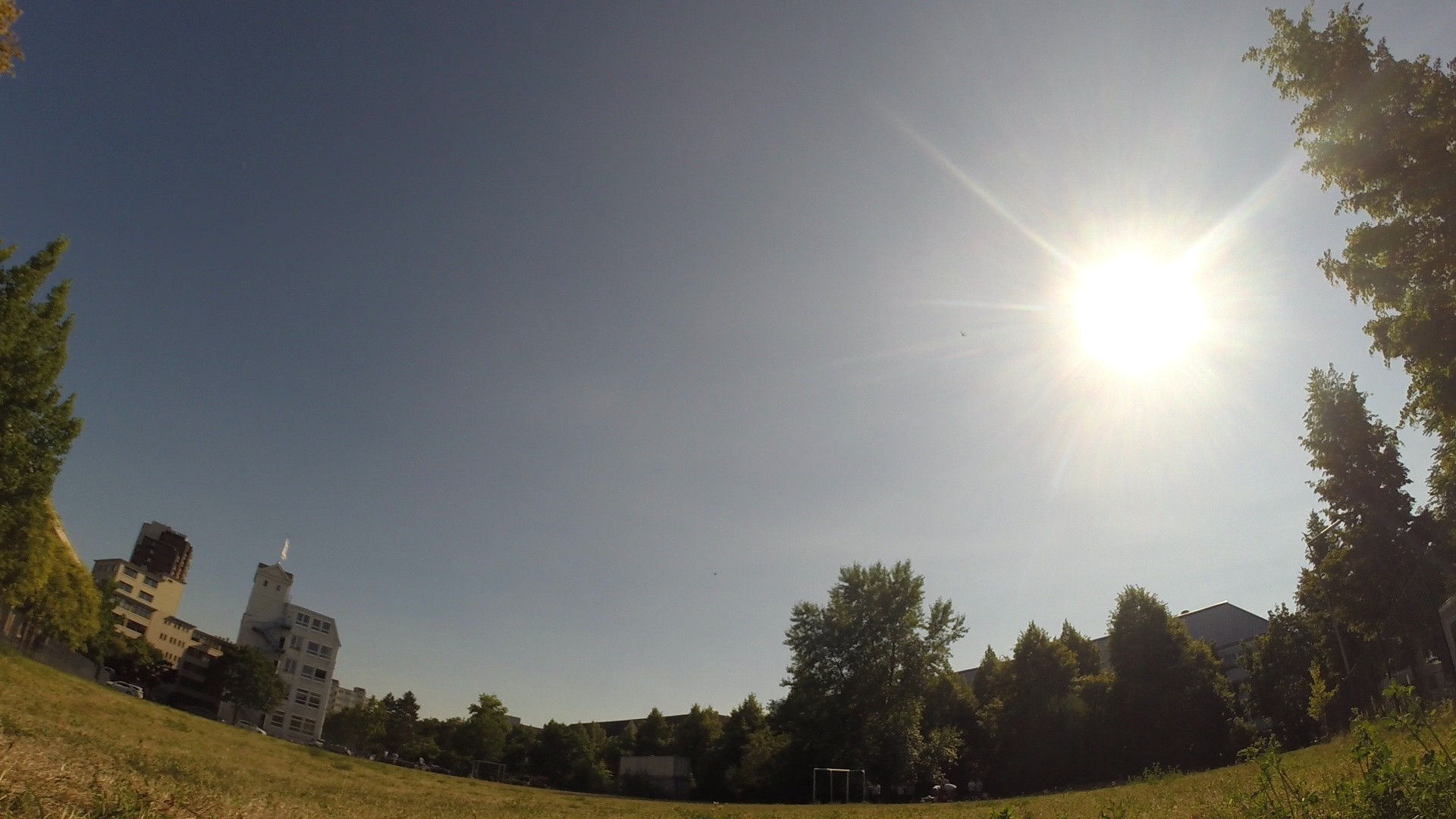}
    }

    \vspace{-2mm}
    
    \caption{Representative images from the \gls*{dds} dataset. Each image contains at least one drone, but bounding boxes were intentionally omitted to emphasize the challenge of visually identifying them in certain~scenarios.}
    \label{fig:samples-dds}
\end{figure}

The competition organizers provided the \gls*{dds}, consisting of 77 annotated video sequences to support method development.
These videos, averaging 1,384 frames each, feature one or more drones, with an average of 1.12 annotated drones per frame. 
Annotations include the frame number and bounding box for each drone.
Although birds frequently appear in the footage, they were left unlabeled.
The dataset exhibits high variability in resolution, ranging from $720\times576$ to $3840\times2160$ pixels.
It also presents sequences with backgrounds of sky or vegetation, diverse weather conditions (cloudy, sunny), direct sun glare, and varying camera characteristics, as shown in \cref{fig:samples-dds}.

Given the absence of predefined guidelines for splitting the \dds dataset into training and validation subsets, a pragmatic, empirically driven approach was adopted.
We initially allocated 70 videos for training and 7 for validation, with a deliberate focus on constructing a relatively diverse validation set.
Specifically, we manually curated the dataset to ensure the validation set included videos with diverse drone distances, image resolutions, backgrounds, and environmental conditions.
Recognizing the potential biases introduced by the initial split, we implemented an iterative refinement process~\cite{laroca2023do}.
This involved strategically swapping videos between the training and validation sets, guided by empirical observations of model performance while maintaining the intended diversity of the validation set.
This methodology, grounded in careful observation and iterative adjustment, aimed to enhance the robustness and generalizability of the trained model.
For reproducibility, the final composition of the validation set is as~follows:
\begin{itemize}
    \item \texttt{2019\_10\_16\_C0003\_3633\_inspire.MP4};
    \item \texttt{dji\_mavick\_mountain.avi};
    \item \texttt{dji\_phantom\_4\_hillside\_cross.avi};
    \item \texttt{GOPR5843\_002.mp4};
    \item \texttt{gopro\_002.mp4};
    \item \texttt{parrot\_disco\_distant\_cross\_3.avi};
    \item \texttt{swarm\_dji\_phantom4\_2.avi}.
\end{itemize}

During our analysis, we uncovered annotation errors within the dataset.
The iterative refinement process of the training and validation splits revealed that some poor results stemmed from annotation inaccuracies rather than detection failures. 
While sporadic errors appeared across several videos, suggesting a semi-automated annotation process, two specific videos, \texttt{2019\_10\_16\_C0003\_3633\_inspire.MP4} and \texttt{two\_distant\_phantom.avi}, exhibited significant inconsistencies.
We hypothesize this resulted from desynchronization between video frames and drone position labels, potentially caused by uncorrected frame deletions at the beginning of the videos.
These errors have since been corrected, and the updated annotations are now publicly~available\customfootnote{\supplementary}.

We employed OpenCV to extract individual frames from the videos, saving them as JPG images with a quality setting of 90.
Although the PNG format would have eliminated compression artifacts, it would have substantially increased the dataset size.
To further optimize processing time and storage, we implemented a frame subsampling strategy, selecting every fifth frame, following~\cite{kim2023high}.
This subsampling process, while potentially introducing a reduction in performance, considerably decreased processing time, thereby enabling a greater number of experimental iterations and a more thorough validation of the annotations, as previously discussed.

\begin{figure}[!htb]
    \centering
    \captionsetup[subfigure]{labelformat=empty,captionskip=2pt}
    
    \resizebox{0.99\linewidth}{!}{
        \subfloat[]{
            \includegraphics[height=9.75ex]{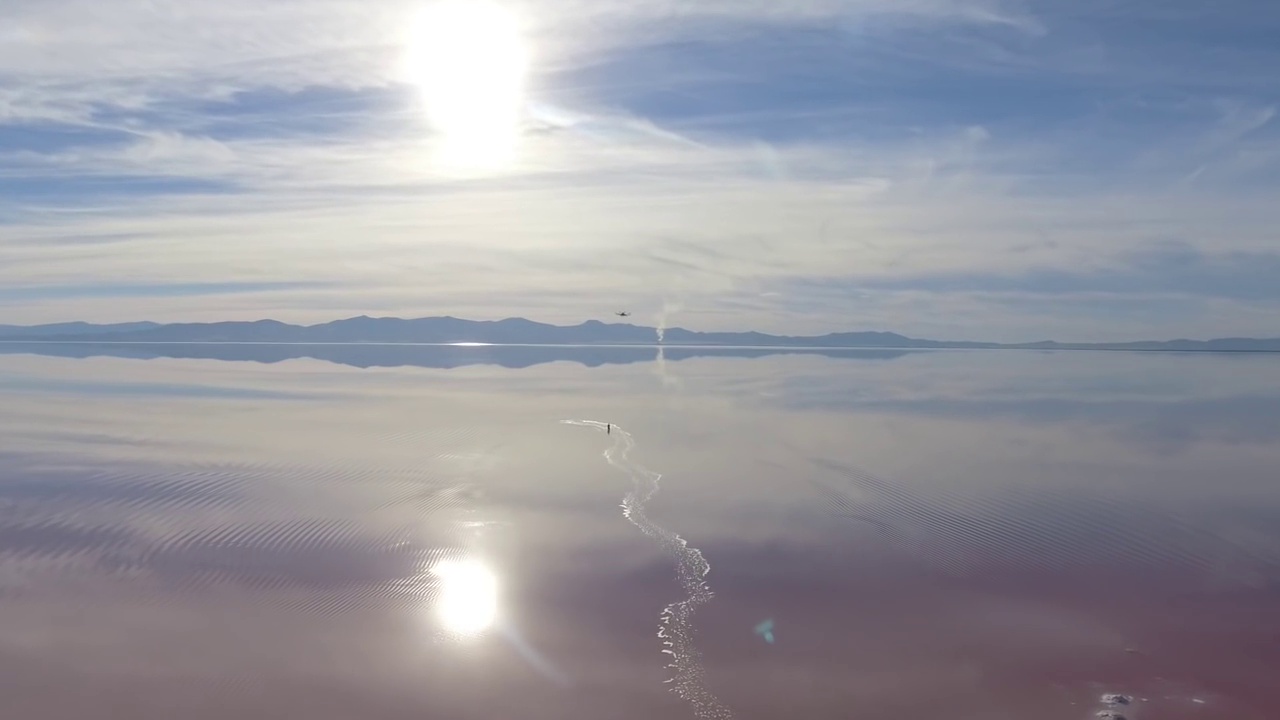}
            \includegraphics[height=9.75ex]{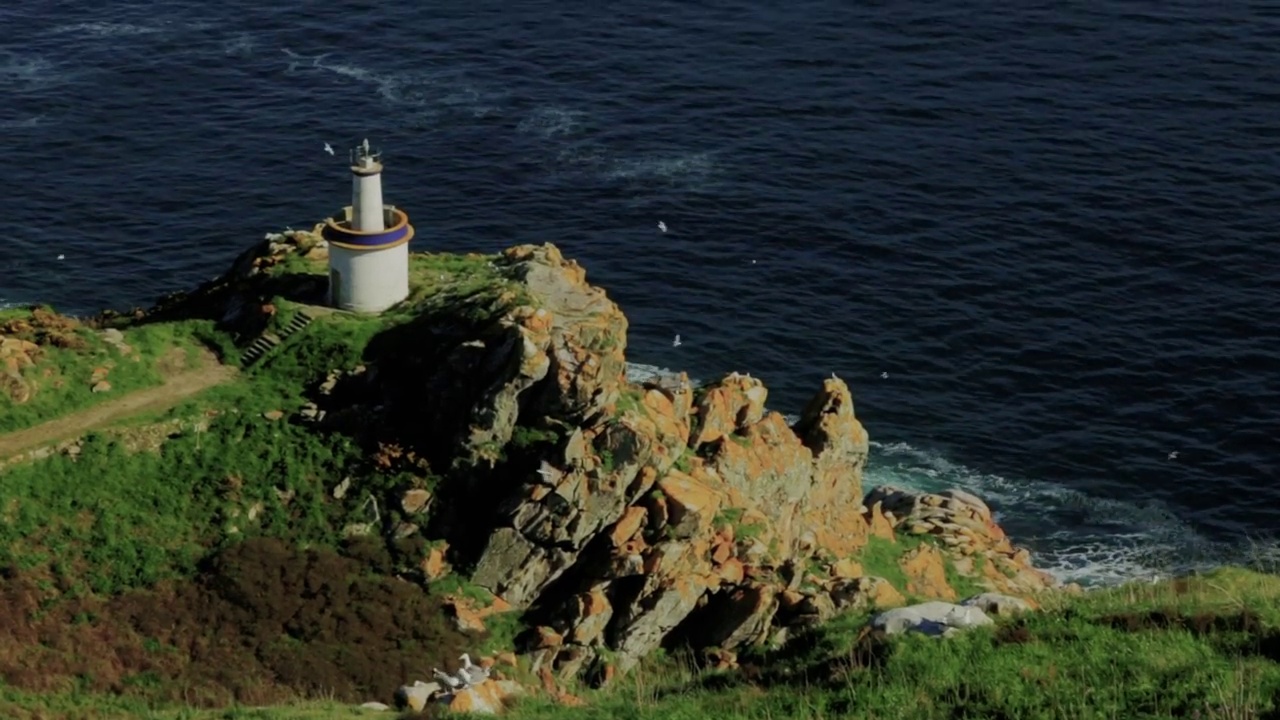}
            \includegraphics[height=9.75ex]{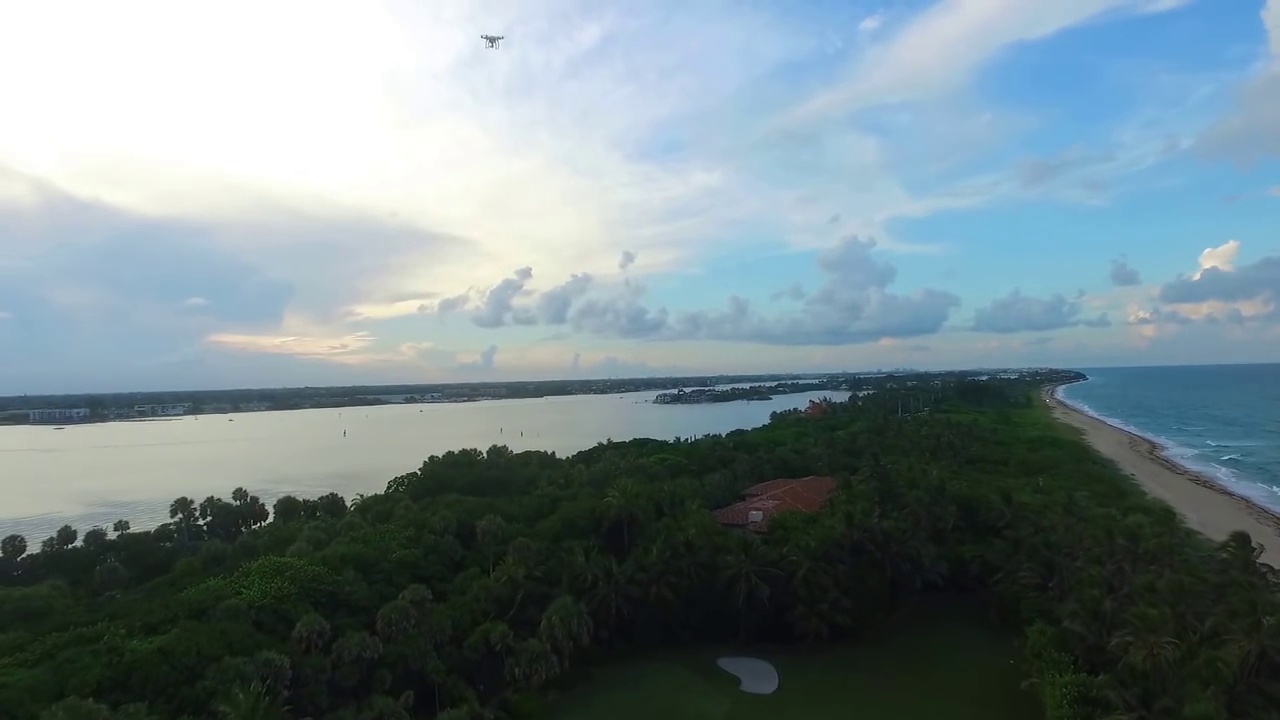}\hspace{1.5mm}
        }
    }

    \vspace{-2.75mm}

    \resizebox{0.99\linewidth}{!}{
        \subfloat[(a)~\gls*{usc}~\cite{bosquet2018stdnet}]{
            \includegraphics[height=9.75ex]{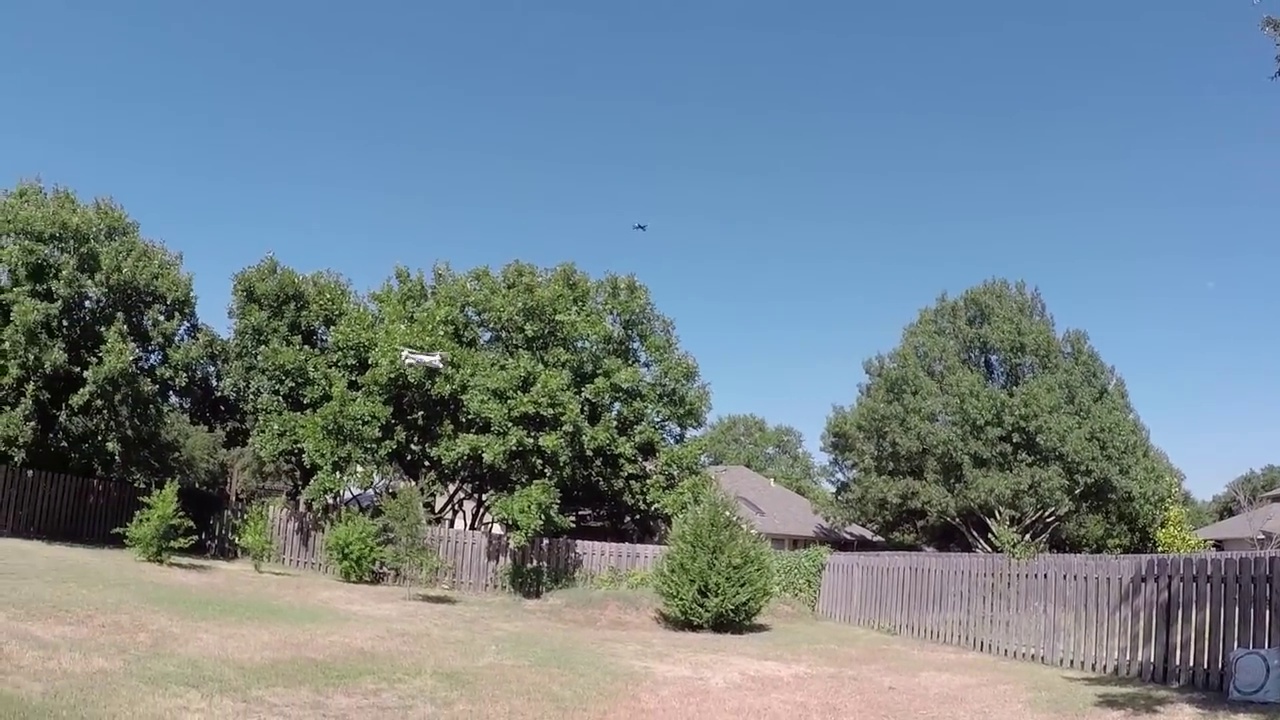}
            \includegraphics[height=9.75ex]{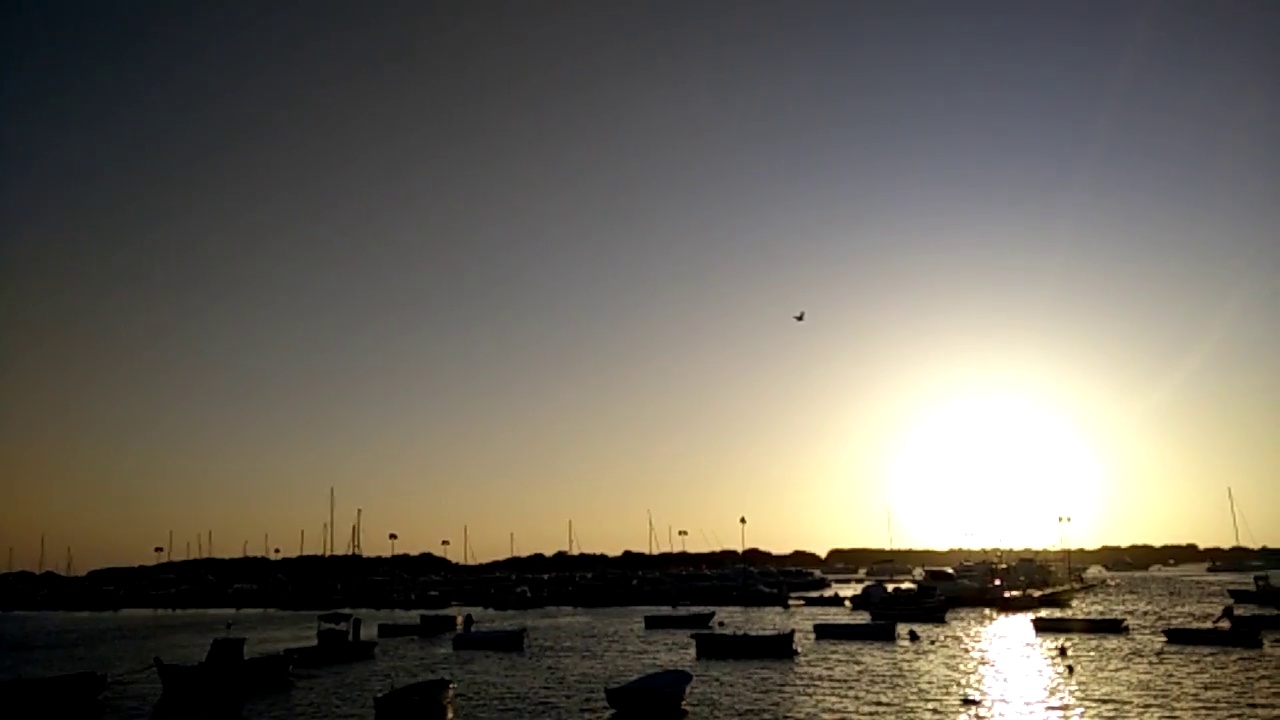}
            \includegraphics[height=9.75ex]{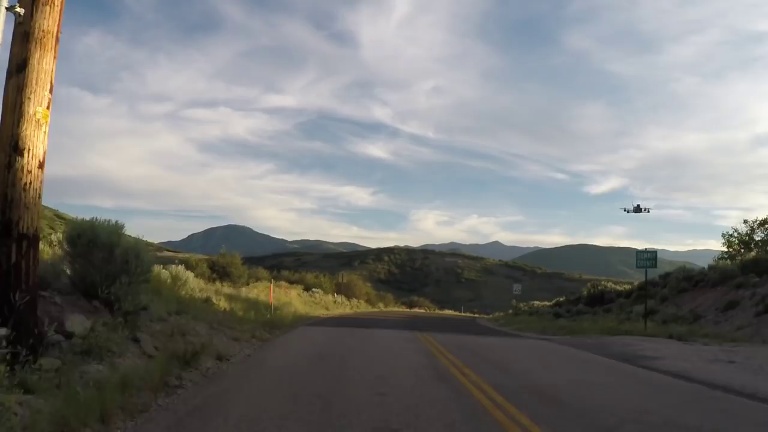}\hspace{1.5mm}
        }
    }

    \vspace{2.5mm}

    \resizebox{0.99\linewidth}{!}{
        \subfloat[]{
            \includegraphics[height=14ex]{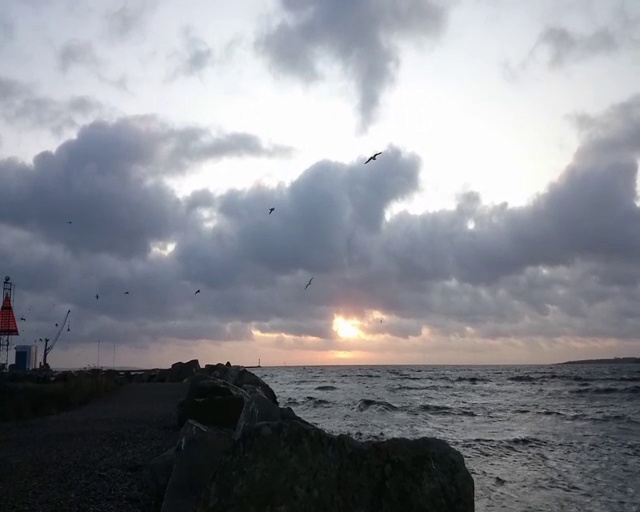}
            \includegraphics[height=14ex]{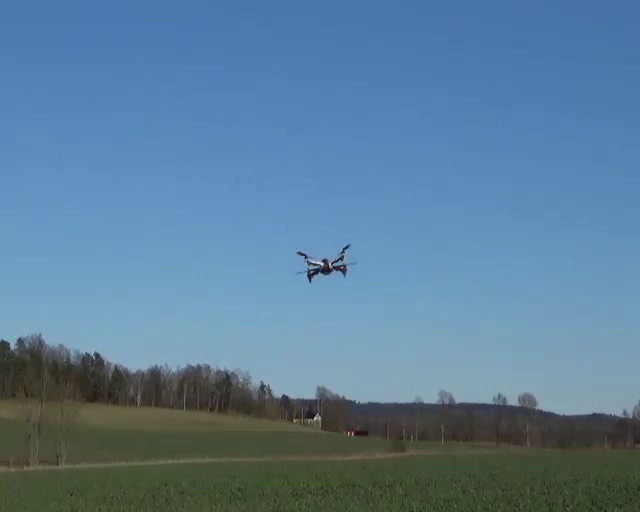}
            \includegraphics[height=14ex]{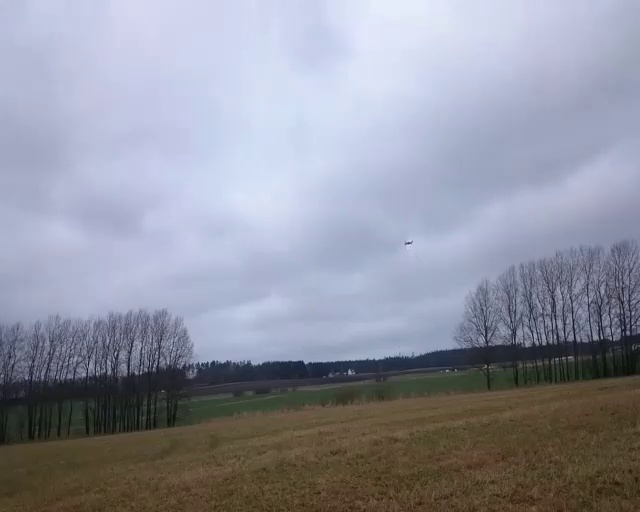}\hspace{1.5mm}
        }
    }
        
    \vspace{-2.75mm}
     
    \resizebox{0.99\linewidth}{!}{  
        \subfloat[(b)~Dataset2~\cite{svanstrom2021real}]{
            \includegraphics[height=14ex]{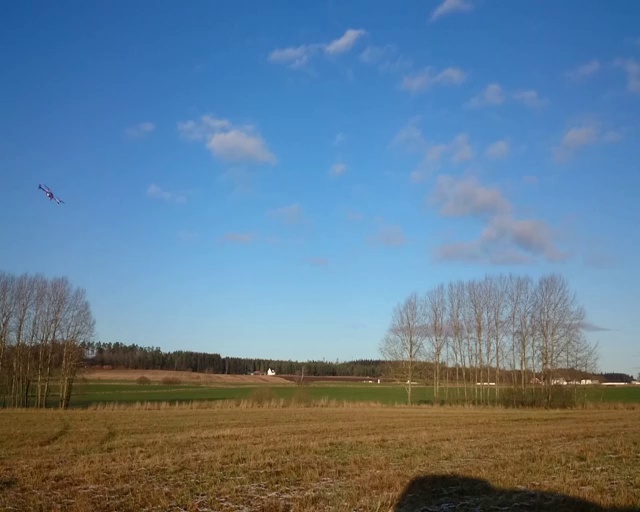}
            \includegraphics[height=14ex]{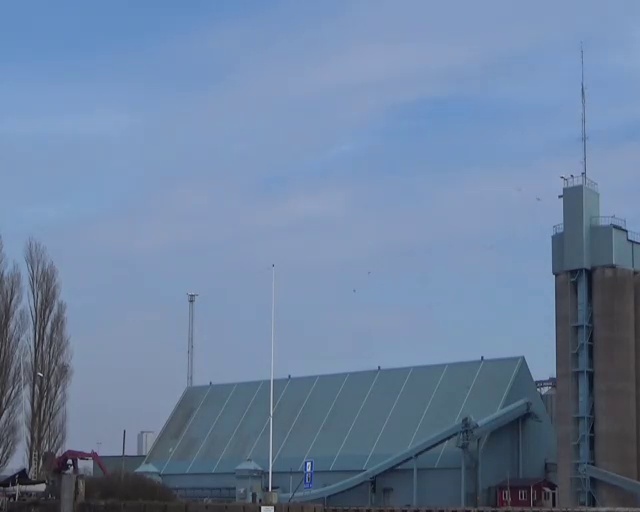}
            \includegraphics[height=14ex]{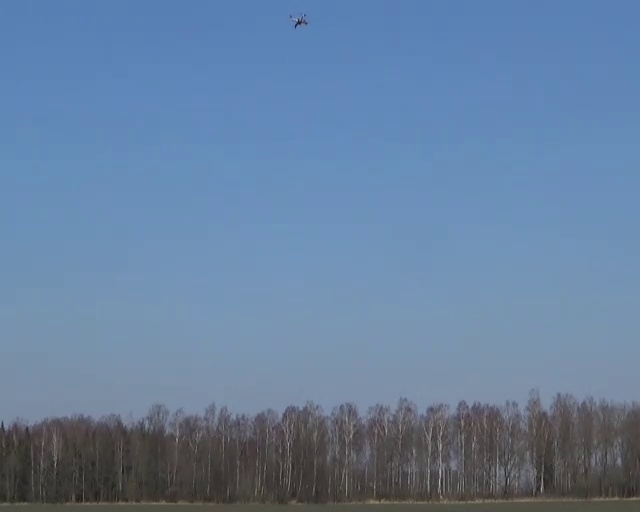}\hspace{1.5mm}
        }
    }

    \vspace{2.5mm}

    \resizebox{0.99\linewidth}{!}{
        \subfloat[]{
            \includegraphics[height=9.75ex]{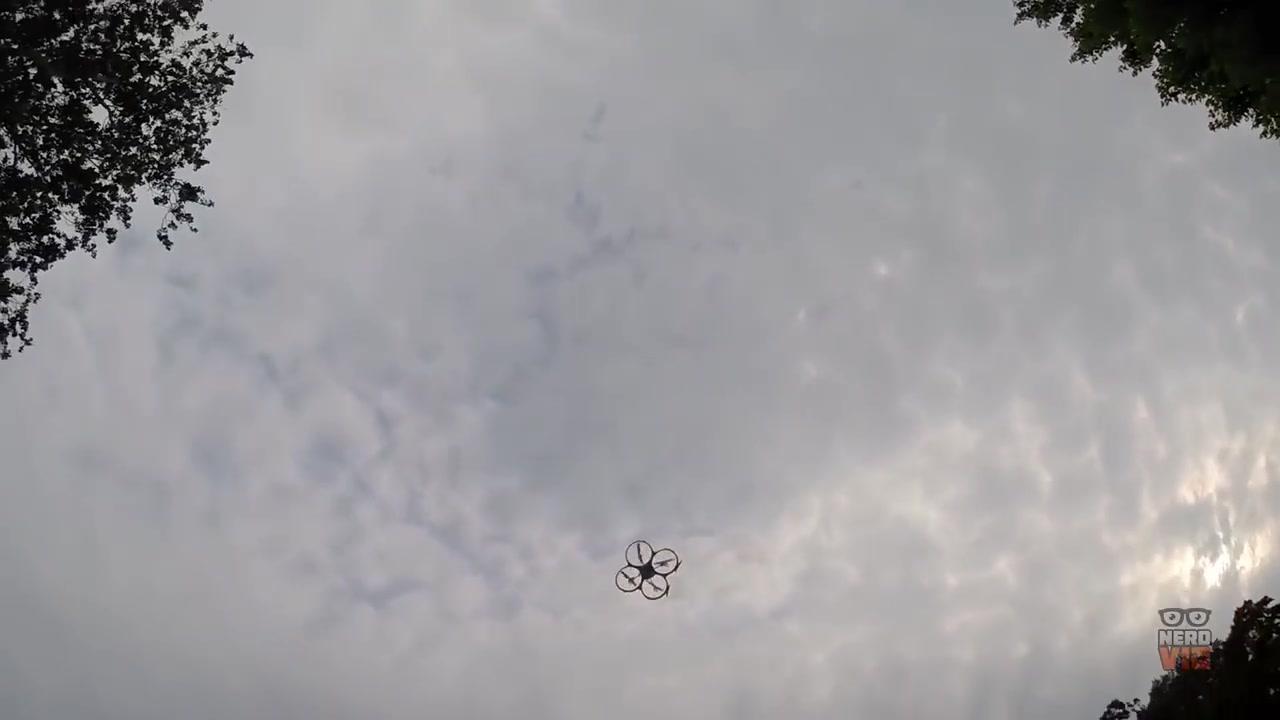}
            \includegraphics[height=9.75ex]{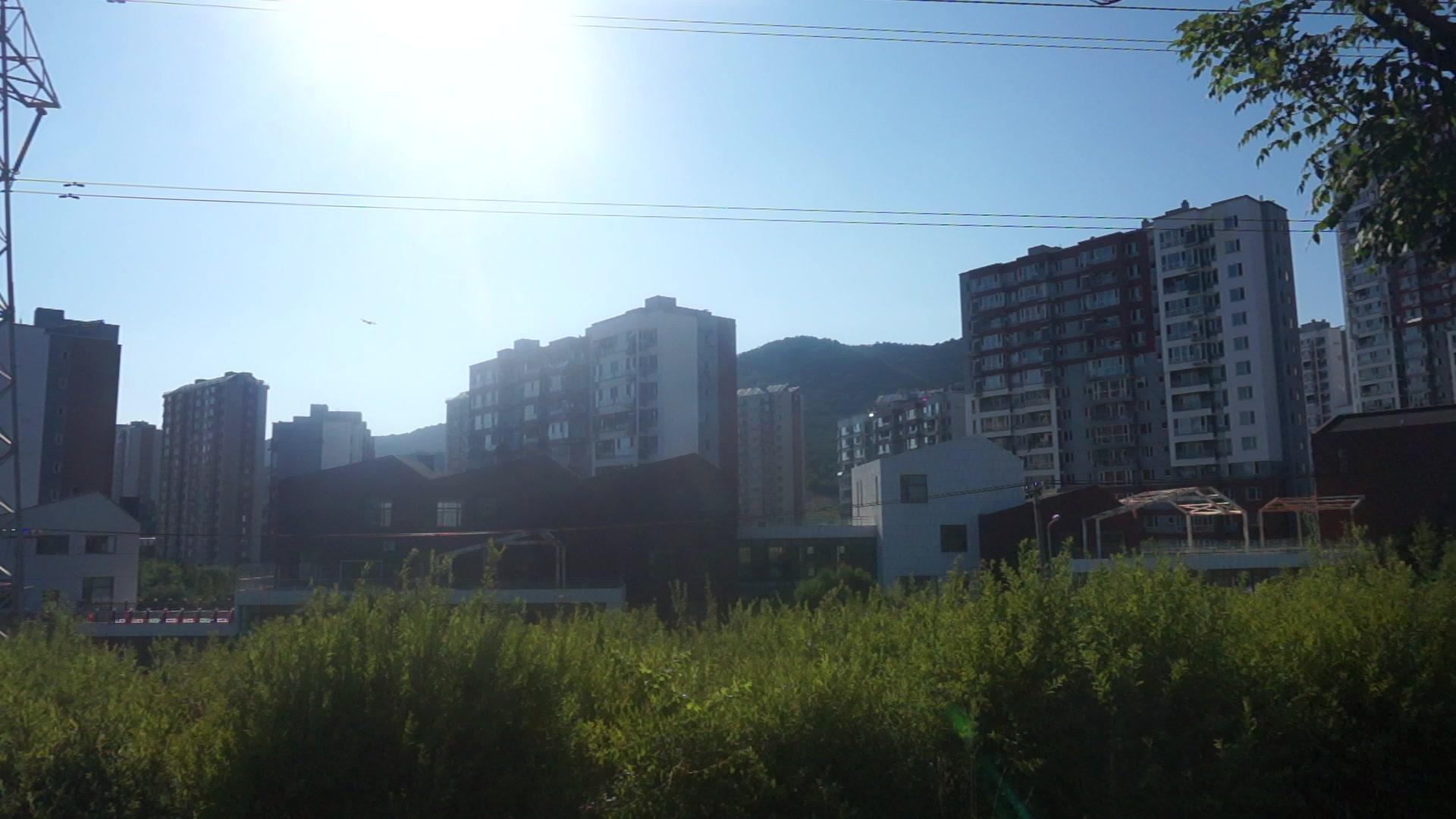}
            \includegraphics[height=9.75ex]{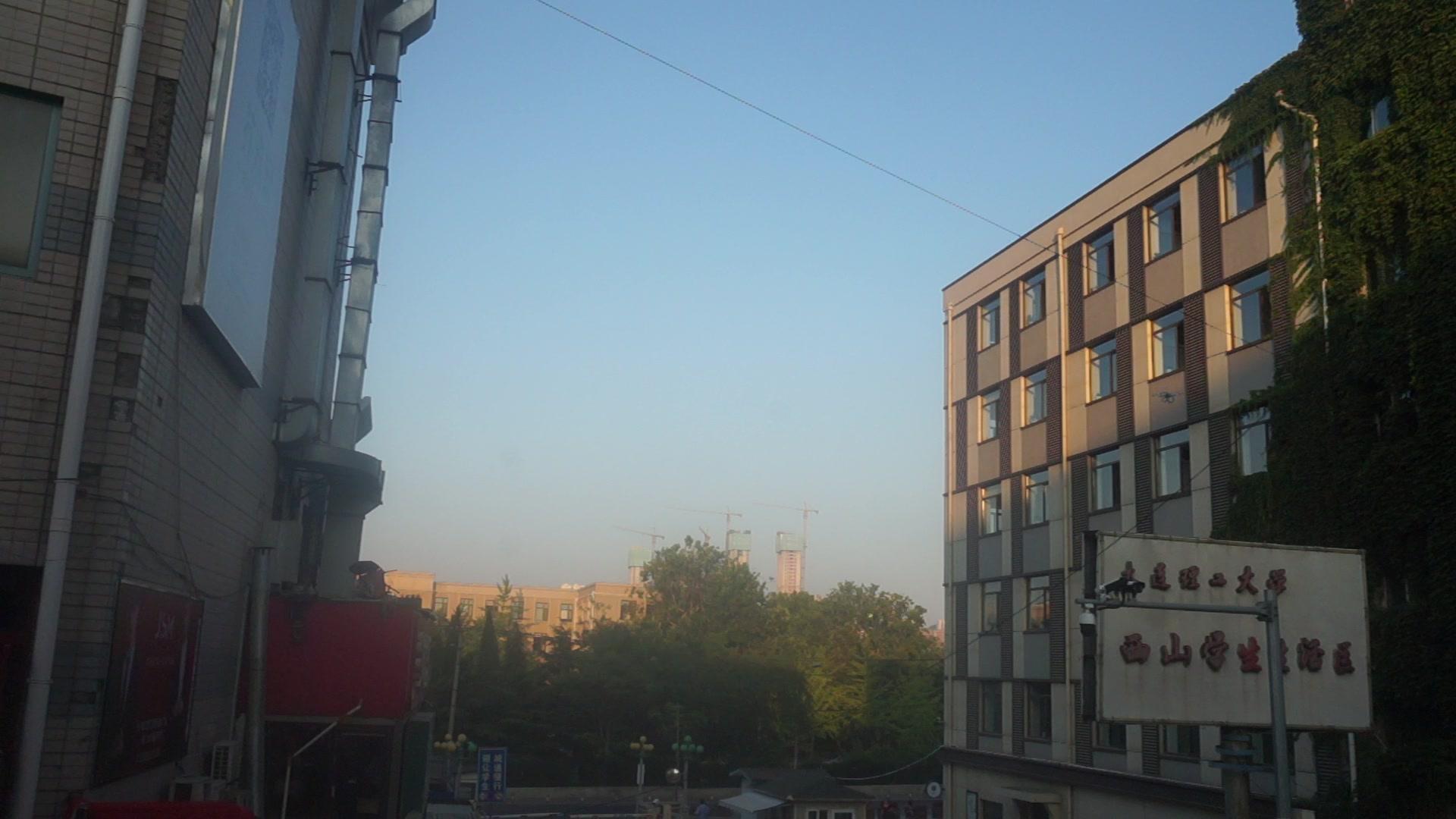}\hspace{1.5mm}
        }
    }

    \vspace{-2.75mm}

    \resizebox{0.99\linewidth}{!}{
        \subfloat[(c)~\dut~\cite{zhao2022vision}]{
            \includegraphics[height=9.75ex]{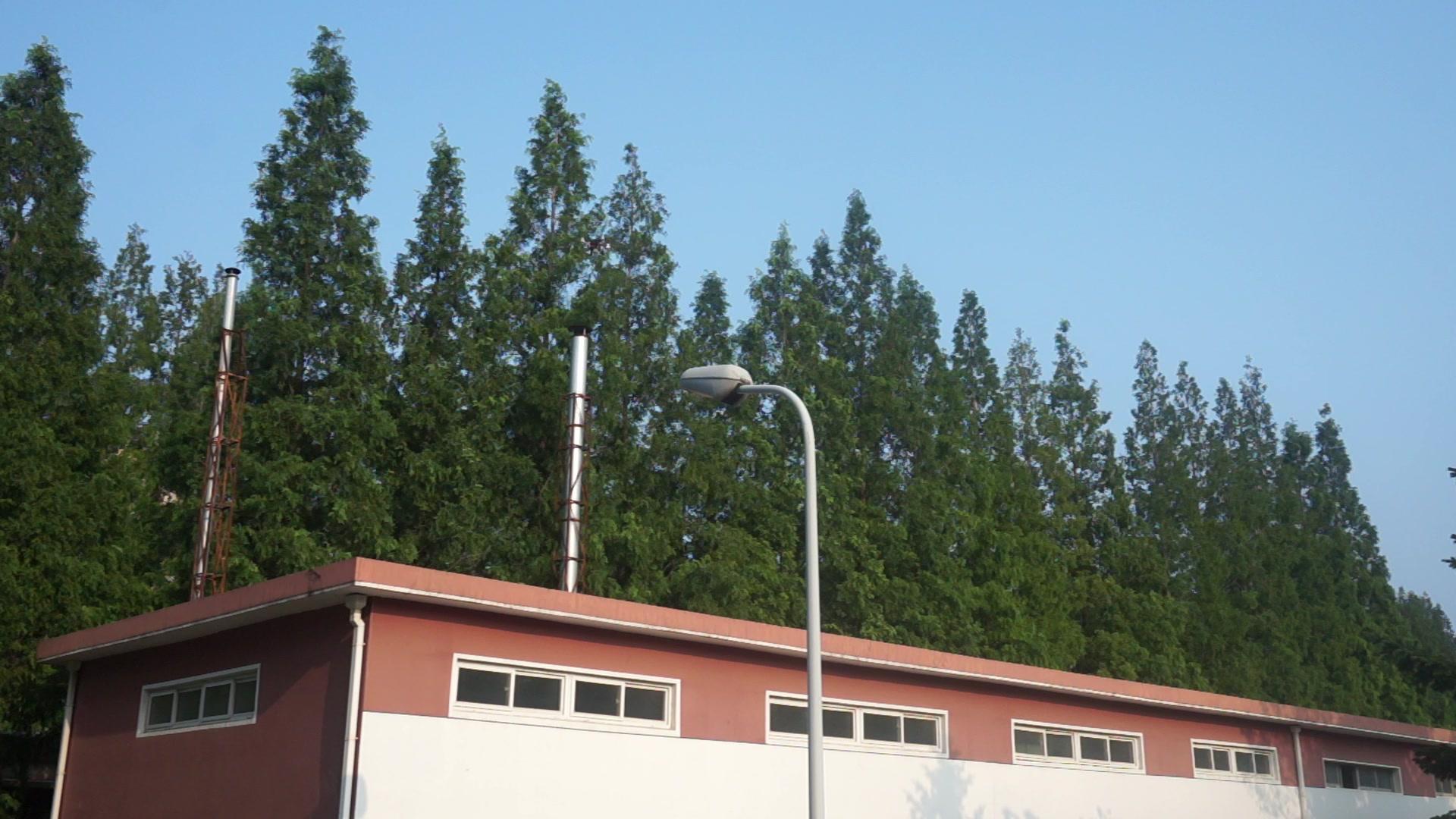}
            \includegraphics[height=9.75ex]{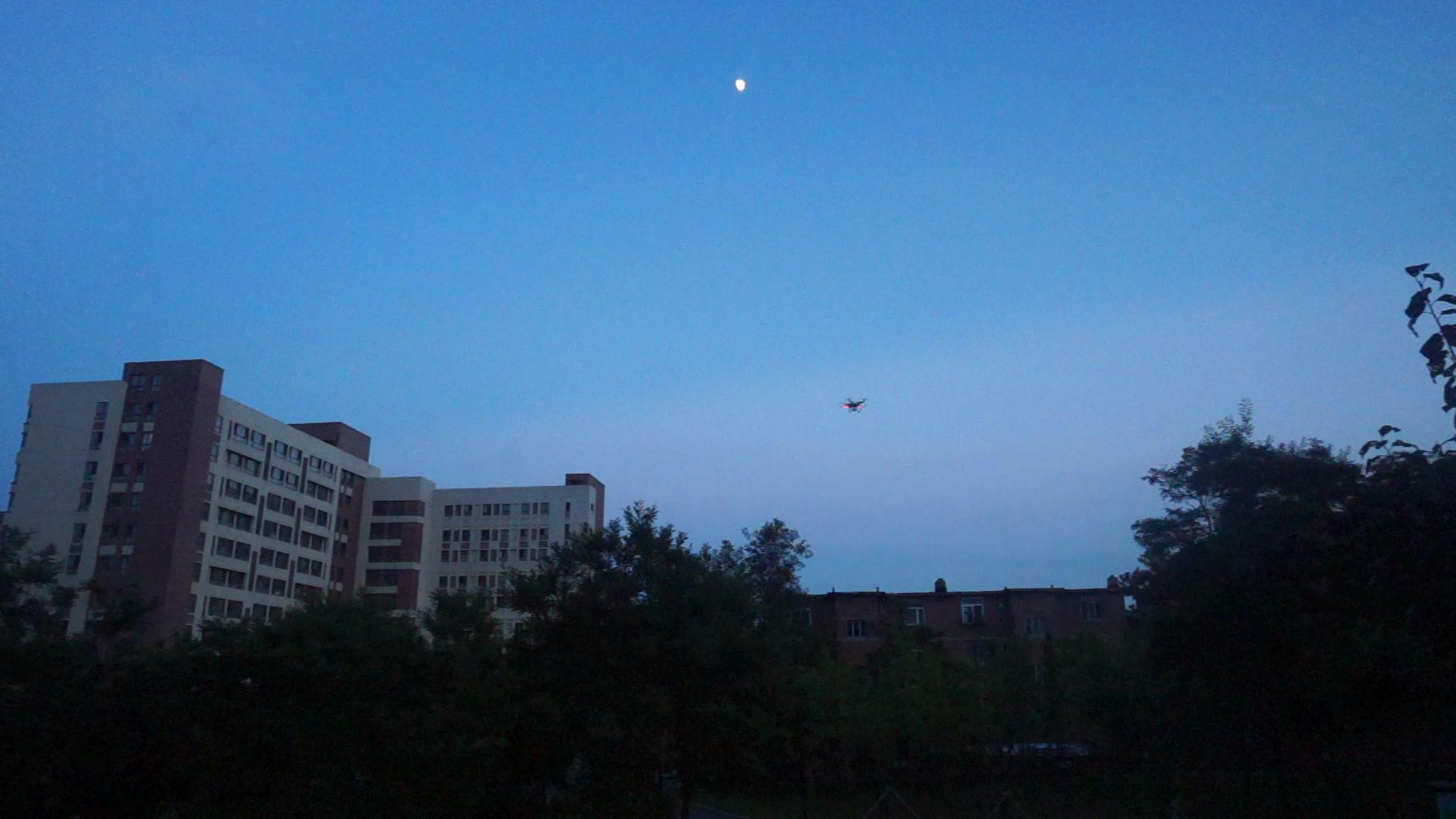}
            \includegraphics[height=9.75ex]{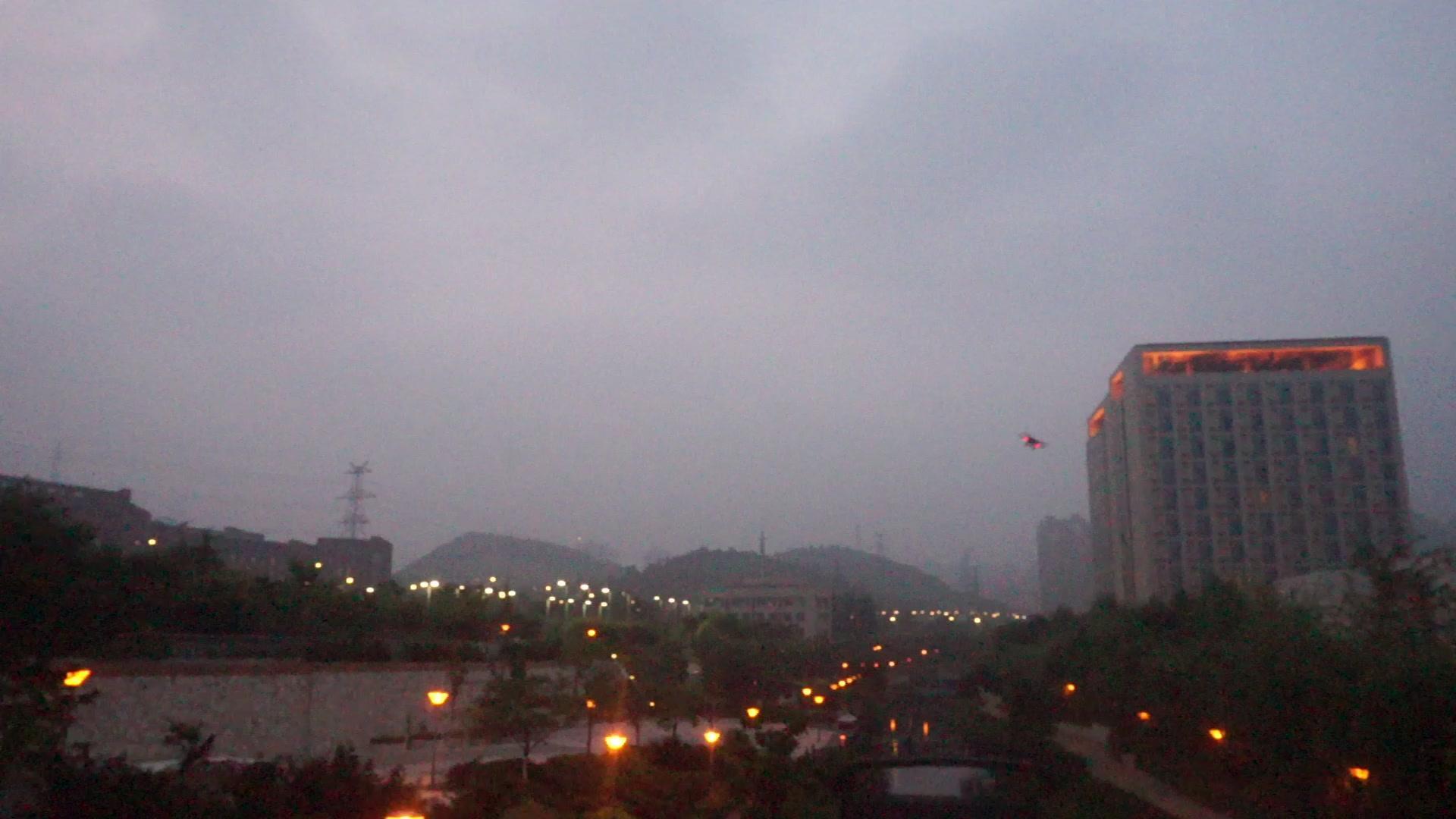}\hspace{1.5mm}
        }
    }
    
    \caption{Samples from the three public datasets integrated into our training~set.}
    \label{fig:samples-public-datasets}
\end{figure}

To improve the robustness of our model across a wider range of scenarios, we incorporated three additional datasets into our training data, as permitted by the competition rules.
These datasets are the \textit{\gls*{usc}}~\cite{bosquet2018stdnet}, \textit{Dataset2}~\cite{svanstrom2021real} (which lacks a formal name in its original publication, hence our designation), and \textit{\dut}~\cite{zhao2022vision}.
A brief description of these datasets, along with representative images, can be found in the subsequent paragraphs and \cref{fig:samples-public-datasets}, respectively.
Note that while these datasets include their own subdivisions for training, testing, and validation, we utilized all available images from them for training our model.
As a result, the validation images consisted exclusively of scenarios from the target~competition.

\gls*{usc}~\cite{bosquet2018stdnet} comprises 115 video segments collected from YouTube, totaling over 25,000 annotated frames.
The videos, captured primarily by drones or from a bird's-eye perspective, have a resolution of approximately $1280\times720$ pixels and cover three main landscapes: air, sea, and land.
The dataset is characterized by small object instances, with pixel areas ranging from 16~($\approx4\times4$) to 256~($\approx 16\times16$) pixels.
Considering our target application, we explored only the 2,263 frames that feature drones or~birds.

Dataset2 contains 365 infrared and 285 visible ten-second videos, totaling 203,328 frames.
We selected 51 visible videos depicting birds and 108 visible videos featuring drones, excluding infrared videos as well as visible videos depicting other objects, such as airplanes and helicopters.
All selected videos have a resolution of $640\times512$ pixels.
A major challenge encountered during the processing of Dataset2 was the provision of annotations in MATLAB format~(.mat), which we could not access due to licensing restrictions.
Efforts to use alternative tools, such as Octave, were unsuccessful and mirrored issues reported by other users on GitHub.
As a result, we resorted to manually selecting and labeling every tenth frame from the selected videos, resulting in 4,516 frames.
These annotations have also been made publicly~available.

The \dut~\cite{zhao2022vision} dataset comprises 10,000 images with 10,109 manually annotated drone positions.
It is distinguished by its high variability, encompassing resolutions from $240\times160$ to $5616\times3744$ pixels, over 35 drone models, and a wide array of backgrounds (sky, clouds, jungles, urban landscapes, farmland, and playgrounds).
Furthermore, it accounts for various lighting conditions (day, night, dawn, and dusk) and weather scenarios (sunny, cloudy, and snowy days).
The drones typically occupy less than 5\% of the image area, though there are cases where the drone nearly covers the entire~image.

While we considered several other datasets from the literature, we ultimately decided not to use them for various reasons.
These included major differences between the datasets and our target scenarios, as well as challenges in accessing the datasets, such as non-functional websites or unresponsive authors regarding the download links, among other~factors.

\subsection{Proposed Approach}
\label{sec:experiments:proposed}

Given the proven track record of YOLO models not only in drone detection~\cite{svanstrom2021real,kim2023high,gao2024novel,munir2024investigation} but also across a broad spectrum of computer vision applications~\cite{laroca2021efficient,diwan2023object,terven2023comprehensive,laroca2023leveraging}, we selected this architecture for our investigation.
Specifically, we leveraged YOLOv11~\cite{yolov11}, a state-of-the-art iteration that builds on the strengths of its predecessors while incorporating enhancements to further improve performance and~flexibility.

This model employs a single-stage, anchor-free architecture optimized for efficiency and ease of deployment.
Its key components include:
(i)~an enhanced backbone that reduces computational costs while maintaining strong feature extraction capabilities;
(ii)~a refined neck structure that improves feature aggregation with fewer parameters, enabling faster inference;
and (iii)~an optimized detection head, designed to reduce latency and enhance speed in the final prediction layers.
For more details on YOLOv11, please refer to~\cite{yolov11}.

YOLOv11 comes in five variants: YOLO11n, YOLO11s, YOLO11m, YOLO11l, and YOLO11x. 
These models vary in size and complexity, directly influencing their speed and accuracy trade-offs.
YOLO11n~(nano) is the smallest and fastest, making it ideal for resource-constrained environments, though with reduced accuracy.
As the models scale up --~YOLO11s~(small), YOLO11m~(medium), YOLO11l~(large), and YOLO11x~(extra-large)~-- they become progressively deeper and wider, enhancing accuracy at the cost of increased computational demands and slower inference speeds.
Ultimately, selecting a variant involves balancing detection performance and speed to meet specific application~requirements.

We started by conducting experiments using the \model model as a baseline.
Nevertheless, we quickly encountered a major challenge: detecting distant drones.
Specifically, when processing high-resolution images, such as 4K or even QHD, and resizing them to the model’s 640-pixel input, distant objects became very small, often approaching or falling below the detection~threshold.

Taking this into account, we proceeded to investigate several alternative models, including but not limited to the variants previously listed in this section, each offering unique configurations and capabilities.
YOLO11m-p2, for example, is a specialized variant with a finer stride configuration optimized for small object detection; however, this comes at the cost of increased computational overhead.
We also experimented with doubling the models' input size (1,280 instead of 640) to capture finer details.
While such variations did lead to some improvements in detection performance, the training times became prohibitively long, and we concluded that the [relatively small] gains did not justify the added computational~cost.

\begin{figure*}[!htb]
    \vspace{-0.5mm}
    
    \centering
    \resizebox{0.99\linewidth}{!}{
    \, \includegraphics[width=0.99\linewidth]{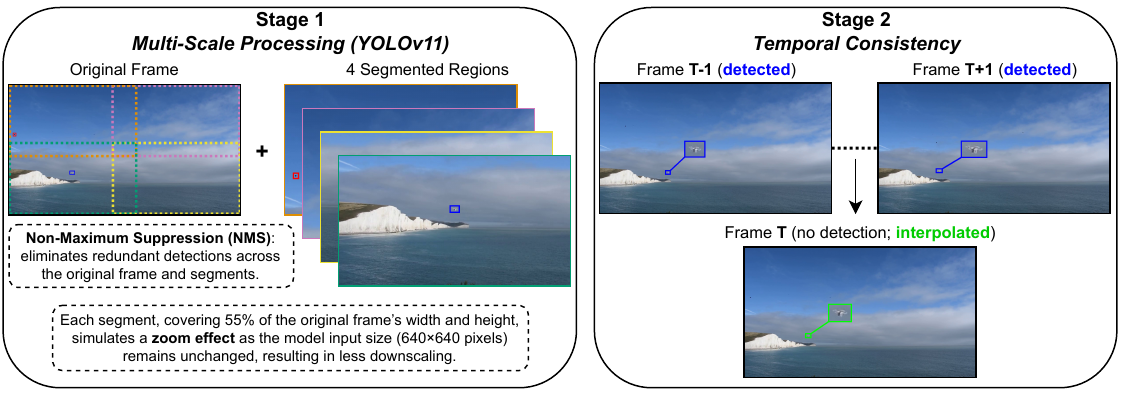}
    }

    \vspace{-4mm}
    
    \caption{Overview of the proposed approach. First, the medium-sized YOLOv11 model (YOLO11m)~\cite{yolov11} is applied to both the full input frame and its segmented regions~(simulating a zoom effect). Detections across the original frame and segments are then aggregated, with redundant bounding boxes removed via \gls*{nms}.
    Lastly, temporal consistency and robustness to missing detections are achieved by tracking drones across a temporal window and applying linear~interpolation.}
    \label{fig:proposed}
\end{figure*}

Recognizing the need to improve detection performance, we decided to process the input images by dividing them into four segments.
Each segment covers 55\% of the original image's width and height, resulting in a small overlap between them.
For example, a $1920\times1080$ image is split into four segments of $1056\times594$ pixels (contrasting with non-overlapping $960\times540$ segments).
To handle drones occupying large portions of the image, the entire image is also processed. 
Finally, after processing both the segments and the full image, non-maximum suppression is applied to eliminate redundant detections.
This strategy, illustrated in \cref{fig:proposed}~(left), yields substantial detection performance gains, as detailed in the next section, thus justifying the increased processing~ overhead.

Based on the preceding analysis, the \model variant was selected over its larger counterparts.
We opted for fine-tuning rather than training from scratch to leverage pre-trained weights and accelerate convergence.
This process was conducted using the well-established Ultralytics framework~\cite{yolov11}, with the following hyperparameter configuration: \gls*{sgd} optimizer with a learning rate of~0.01 and a momentum of~0.9, a batch size of~16, an early stopping patience value of~15, and an \gls*{iou} threshold for \gls*{nms} of~0.1.

One of the core challenges of the competition lies in accurately distinguishing drones from birds, hence its name.
In this sense, we trained our model to detect two distinct classes: drone and bird.
Although the \dds dataset contains images with birds, it does not provide corresponding labels.
In contrast, the \usc dataset, which we incorporated into our training data, includes annotations for both classes.
During inference, however, we focused exclusively on drone detection, ignoring predictions for the bird class.
We also experimented with training a model solely for drone detection (i.e., without the bird class), but this approach resulted in slightly worse~results.

To further enhance performance, we extensively applied data augmentation.
\cref{tab:data-aug-parameters} details the specific parameters used for each transformation, adhering to the naming conventions of the Ultralytics framework.
These transformations, standard practice in object detection, introduce variability into the training data, thereby improving model generalization.
We conducted multiple iterations with varying parameter settings to determine the optimal values, evaluating their impact empirically.
For instance, while the default values for scale and perspective transformations are $0.5$ and $0$, respectively, we found that $0.3$ and $0.0001$ yielded better results in our target scenario.
For an explanation of each parameter and its specific effects, refer to the official Ultralytics documentation\customfootnote{\url{https://docs.ultralytics.com/usage/cfg/}}.

\begin{table}[!htb]
\centering
\caption{Data augmentation parameters, following the naming conventions employed in the Ultralytics framework.}
\label{tab:data-aug-parameters}

\vspace{-2mm}

\begin{tabular}{lc}
    \toprule
    Parameter & Value \\ \midrule
    bgr & $0.05$ \\
    degrees & $10$ \\
    erasing & $0.15$ \\
    flipud & $0.0$ \\
    fliplr & $0.5$ \\
    hsv\_s & $0.7$ \\ \bottomrule
\end{tabular}
\,
\begin{tabular}{lc}
    \toprule
    Parameter & Value \\ \midrule
    hsv\_v & $0.4$ \\
    mosaic & $0.1$ \\
    perspective & $0.0001$ \\
    scale & $0.3$ \\
    shear & $10$ \\
    translate & $0.1$ \\
    \bottomrule
\end{tabular}
\end{table}

Beyond standard transformations, we applied a copy-paste technique to improve the training set with additional drone and bird instances.
This approach involved randomly selecting patches of drones and birds and placing them in new, randomly chosen locations within the existing training images.
This process is detailed in the following~paragraph.

\begin{figure}[!htb]
    \centering
    \includegraphics[width=0.45\linewidth]{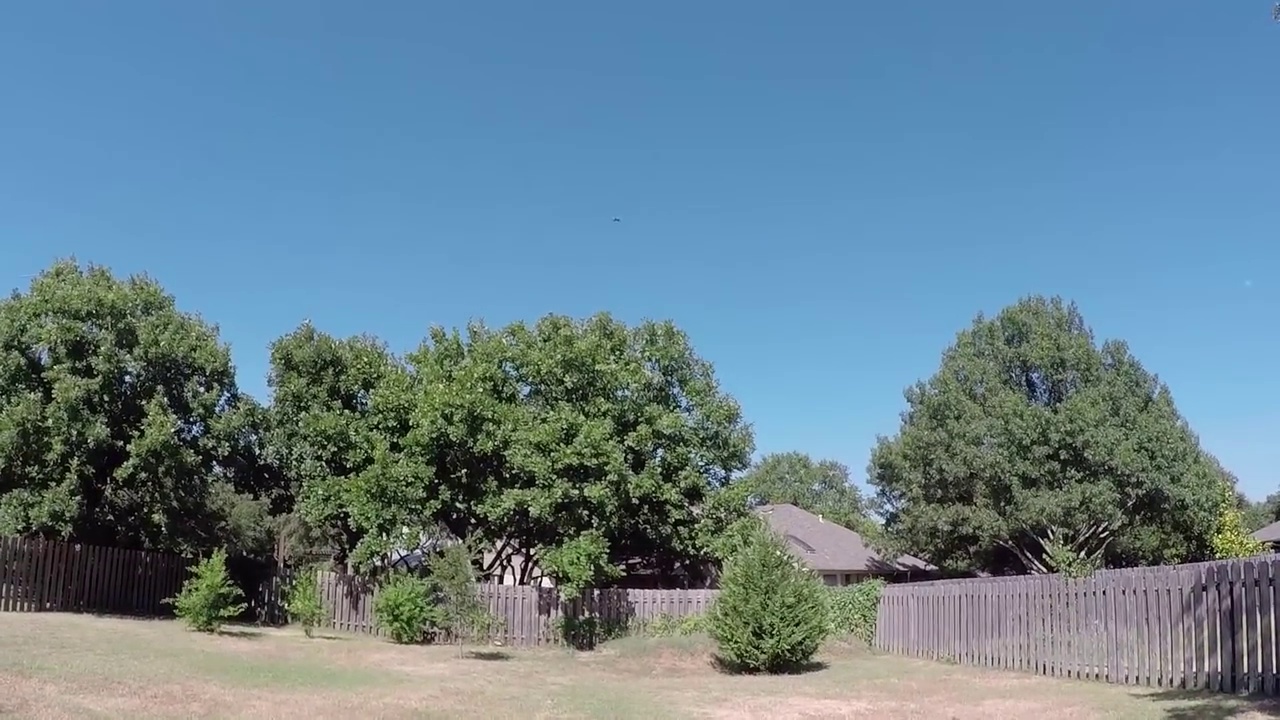}
    \includegraphics[width=0.45\linewidth]{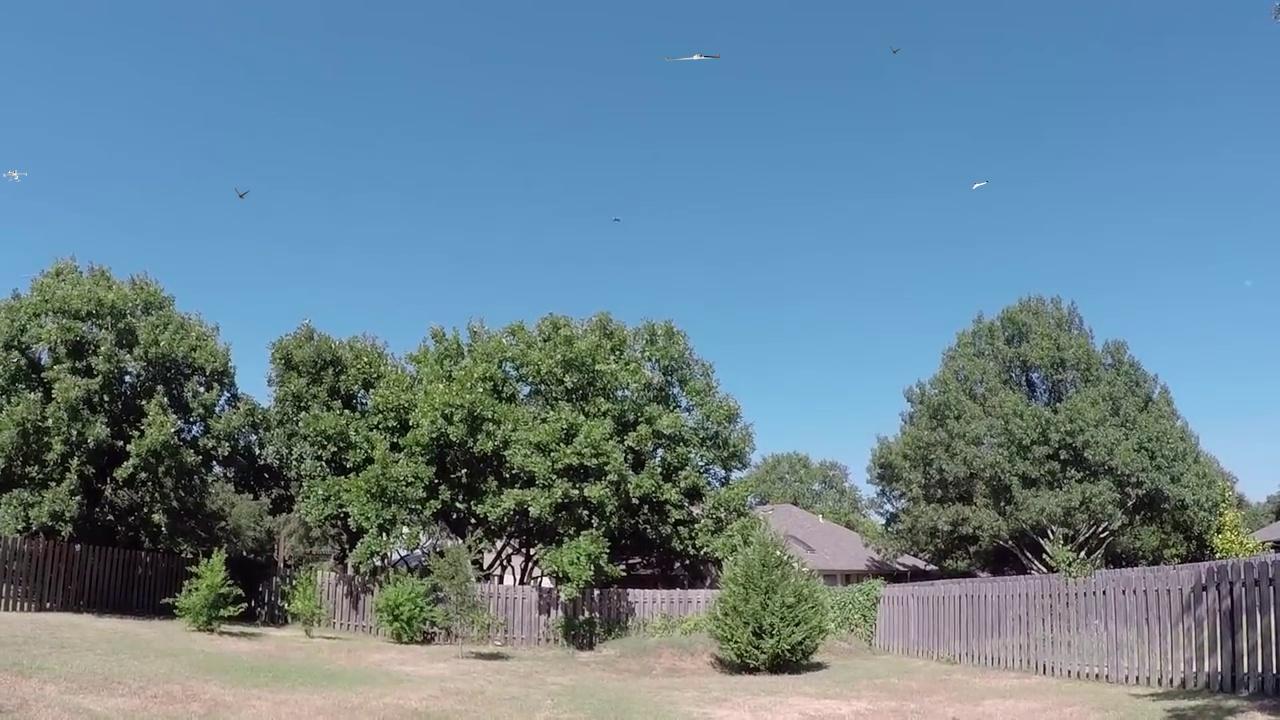}

    \vspace{1mm}
    
    \includegraphics[width=0.45\linewidth]{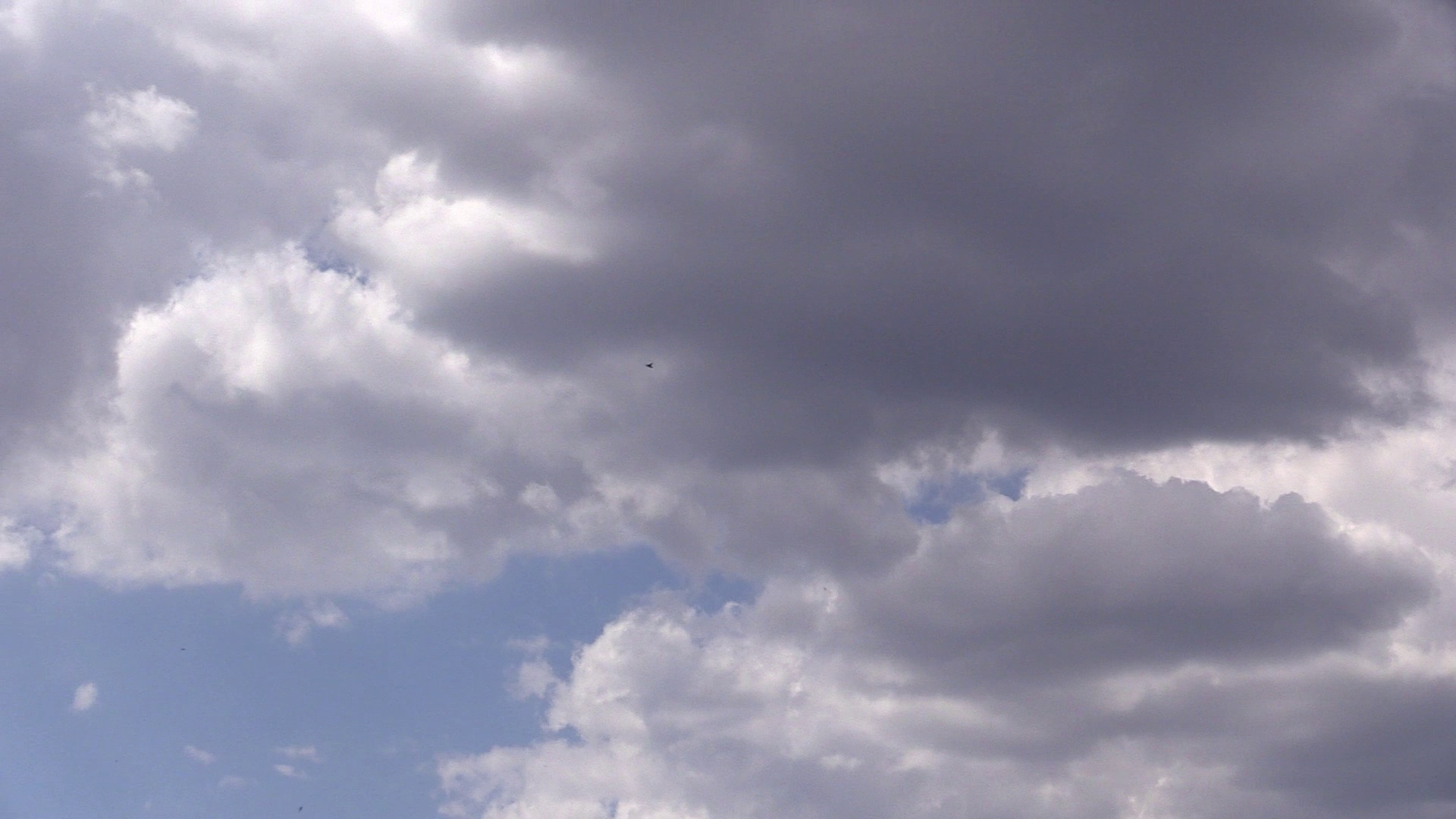}
    \includegraphics[width=0.45\linewidth]{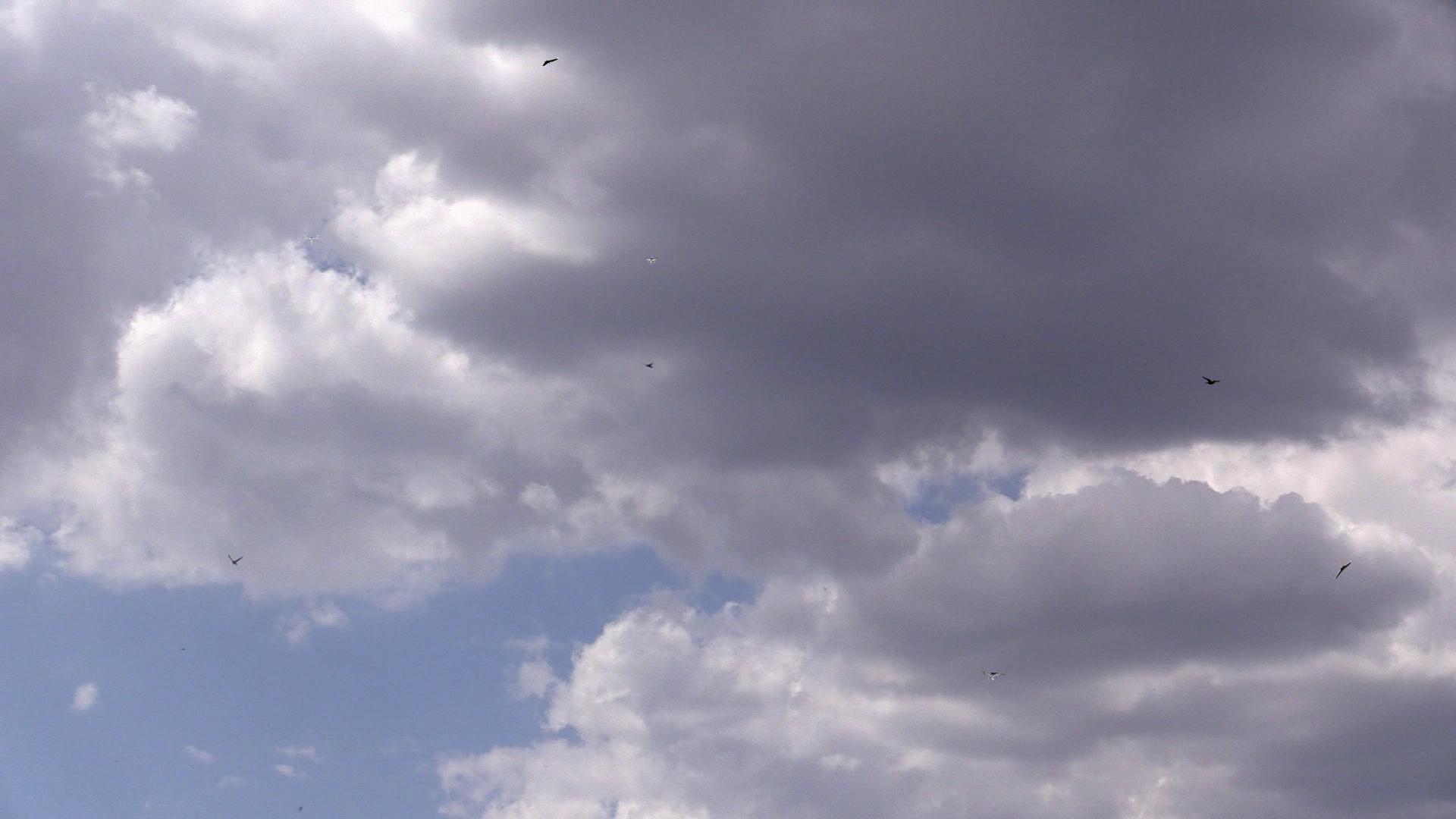}

    \vspace{-2mm}
    
    \caption{Demonstration of the applied copy-paste data augmentation technique: original images~(left) and their augmented counterparts~(right), where patches of drones and birds were inserted to enhance variability in the training~set.}
    \label{fig:samples-copy-paste}
\end{figure}

The images used for pasting were collected from both the training set and various online sources, all featuring transparent backgrounds to ensure seamless integration.
To enhance variability, these patches were augmented using the Albumentations library~\cite{albumentations}, applying transformations such as blurring, pixel dropout, noise addition, and adjustments to brightness, contrast and gamma.
Each patch instance was resized based on randomly selected scale factors within predefined ranges and positioned realistically within the image.
Overlap checks and color difference~(\(\Delta E\)) evaluations were performed to maintain visual consistency.
Once a suitable insertion location was identified, the augmented instance was incorporated into the image, and its corresponding bounding box information was stored.
\cref{fig:samples-copy-paste} provides a visual representation of this~process.

To optimize detection performance, a confidence threshold of~0.375 was experimentally established.
To mitigate missed detections, a post-processing stage was implemented.
Specifically, this routine examines each frame's detected objects and searches for matching objects within a temporal window spanning six~frames both preceding and following the target frame.
The matching criteria require objects to share the same classification label and have a minimum degree of overlap~(\gls*{iou}~$\ge$~0.1).
Upon identifying a potential match, the bounding box of the interpolated object is computed by linearly interpolating the bounding box coordinates between the matched objects.
To ensure reliability, objects near image boundaries are discarded to prevent extrapolating detections for objects that have left the frame.
Furthermore, the interpolated object is accepted only if it does not significantly overlap with existing detections in the target frame, avoiding redundant predictions.
The confidence score of the interpolated object is computed as the mean confidence of the matching objects, reduced by a factor of~2 to account for the uncertainty introduced by interpolation.
As illustrated in \cref{fig:proposed}~(right), this process helps bridge detection gaps, improving temporal consistency and overall accuracy, albeit with a minor increase in the number of false~positives.

Another post-processing strategy we explored involved incorporating a classifier to refine the regions detected by \model, aiming to further reduce false positives.
We experimented with various models, ranging from lightweight networks designed for specific tasks~\cite{laroca2021towards,laroca2022first} to well-established architectures such as \mobilenet~\cite{howard2019searching}.
To minimize the impact on inference time, we prioritized smaller models.
However, regardless of the model used, performance declined.
While this approach effectively filtered out many false positives, it also resulted in the loss of many true positives (in other words, several drones correctly detected by \model were erroneously discarded by the~classifier).

We chose not to implement a maximum-drone-count heuristic, as utilized in~\cite{mistry2023drone}, due to its limited real-world applicability.
While such heuristics can be effective in competition environments with short, pre-defined videos, they struggle to adapt to real-time video streams' dynamic and unpredictable~nature.

\subsection{Results}
\label{sec:experiments:results}

We evaluated our approach on the validation set, which consists of seven videos from the Drone-vs-Bird challenge dataset, covering diverse conditions as described in \cref{sec:experiments:datasets} (the challenge organizers maintained exclusive access to the test set annotations).
Performance was assessed using the \gls*{map} at an \gls*{iou} threshold of 50\%~(\gls*{map}$_{50}$).
This metric calculates the average precision across recall values, considering a detection successful if the \gls*{iou} between the predicted and ground truth bounding boxes is at least~50\%.

The results, presented in \cref{tab:results}, show the \gls*{map}$_{50}$ achieved for each video.
We compare two approaches: our full proposed method, which employs a multi-scale input processing strategy analyzing both whole images and segmented components, and a simplified variant processing only whole~images.

\begin{table}[!htb]
    \centering
    \caption{Results obtained for each video in the validation~set.}
    \label{tab:results}
    
    \vspace{-2mm}
    
    \begin{tabular}{lcc}
    \toprule
    Video Name   & mAP$_{50}$ & mAP$_{50}$$^\dagger$    \\ \midrule
    \texttt{dji\_mavick\_mountain}        & $0.9891$ & $0.6431$ \\
    \texttt{2019\_10\_16\_C0003\_3633\_inspire}        & $0.9421$ & $0.9219$ \\
    \texttt{parrot\_disco\_distant\_cross\_3}        & $0.8684$ & $0.5550$ \\
    \texttt{GOPR5843\_002}        & $0.7175$ & $0.3371$ \\
    \texttt{swarm\_dji\_phantom4\_2}        & $0.7077$ & $0.6566$ \\
    \texttt{dji\_phantom\_4\_hillside\_cross}        & $0.4992$ & $0.7406$ \\
    \texttt{gopro\_002}        & $0.4491$ & $0.0121$ \\ \midrule
    Average & $0.7390$ & $0.5523$ \\ \bottomrule
    \noalign{\vskip 0.4ex}
    \multicolumn{3}{l}{\hspace{-2mm}$^\dagger$\hspace{0.3mm}\scriptsize{using the simplified variant that processes only whole images.}}
    \end{tabular}%
\end{table}

\cref{tab:results} clearly shows that processing the input images both as a whole and in segmented components improves detection performance, particularly for smaller drones. However, the \texttt{dji\_phantom\_4\_hillside\_cross} video stands as an exception, where processing only whole images resulted in better outcomes.
Upon further analysis, we determined that the reason for this was the video's complex hillside background, characterized by distant rocks and vegetation that closely mimic small-scale drones.
When examining cropped bounding boxes from these regions, we occasionally mistook them for drones as well.
In this particular video, the drone is white, providing a strong contrast against the background, and is relatively large.
As a result, dividing the input image into segments was not strictly necessary.
While incorporating segmented components in the processing still improved Recall, it resulted in a disproportionately larger decrease in~Precision.

\begin{figure}[!htb]
    \centering

    \resizebox{0.99\linewidth}{!}{
        \includegraphics[height=14ex]{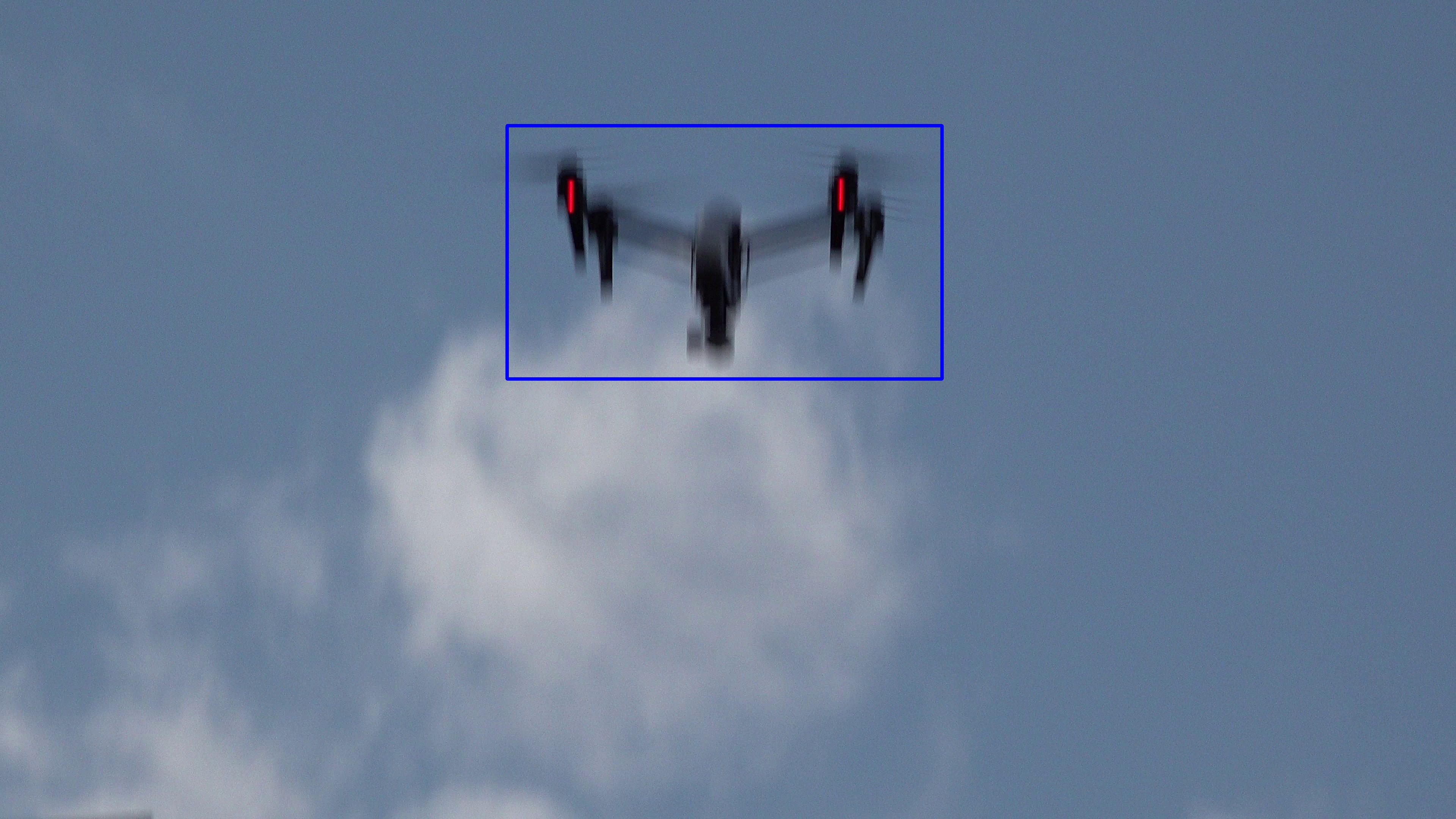}
        \includegraphics[height=14ex]{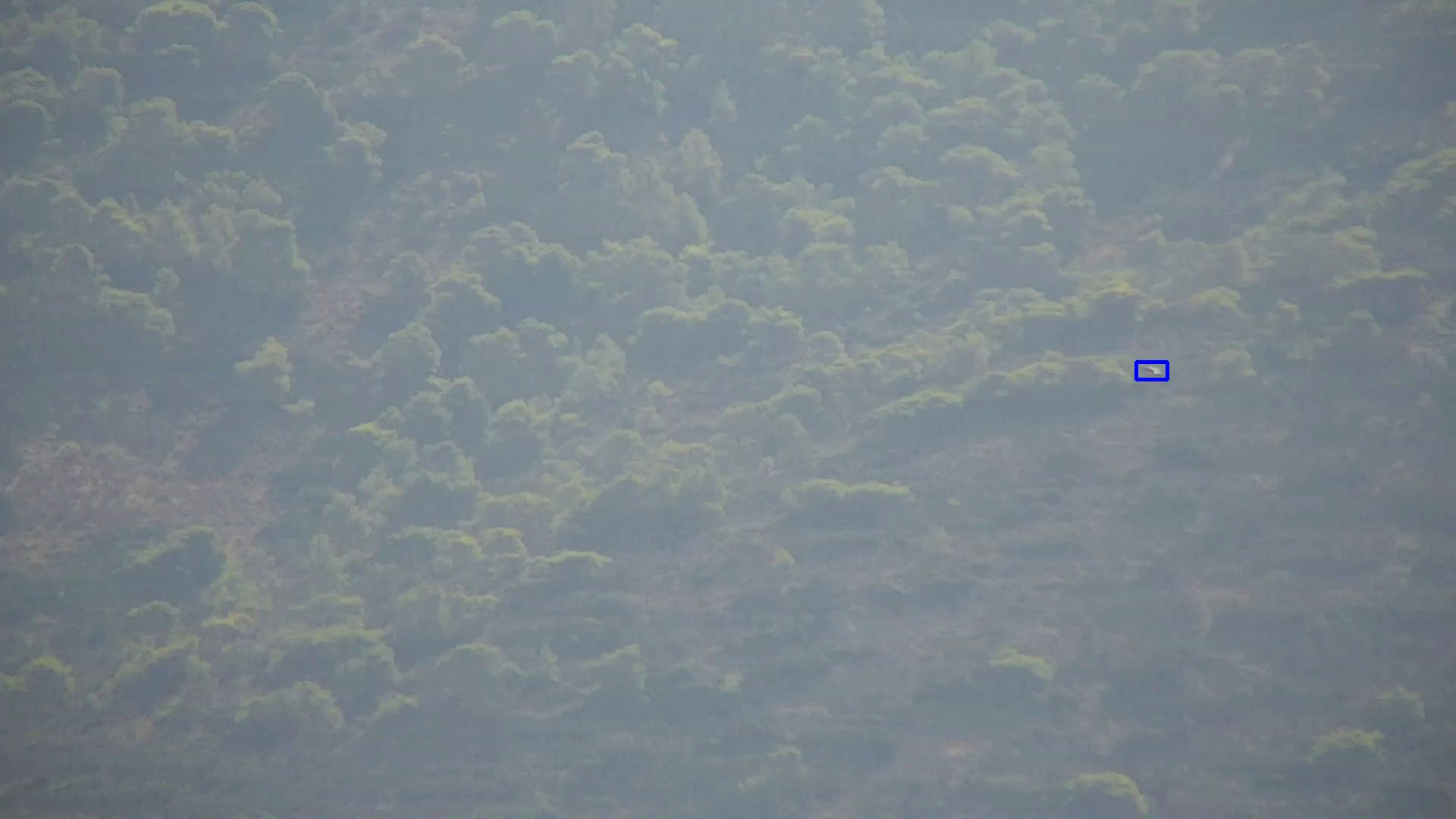}
    }

    \vspace{1mm}

    \resizebox{0.99\linewidth}{!}{
        \includegraphics[height=14ex]{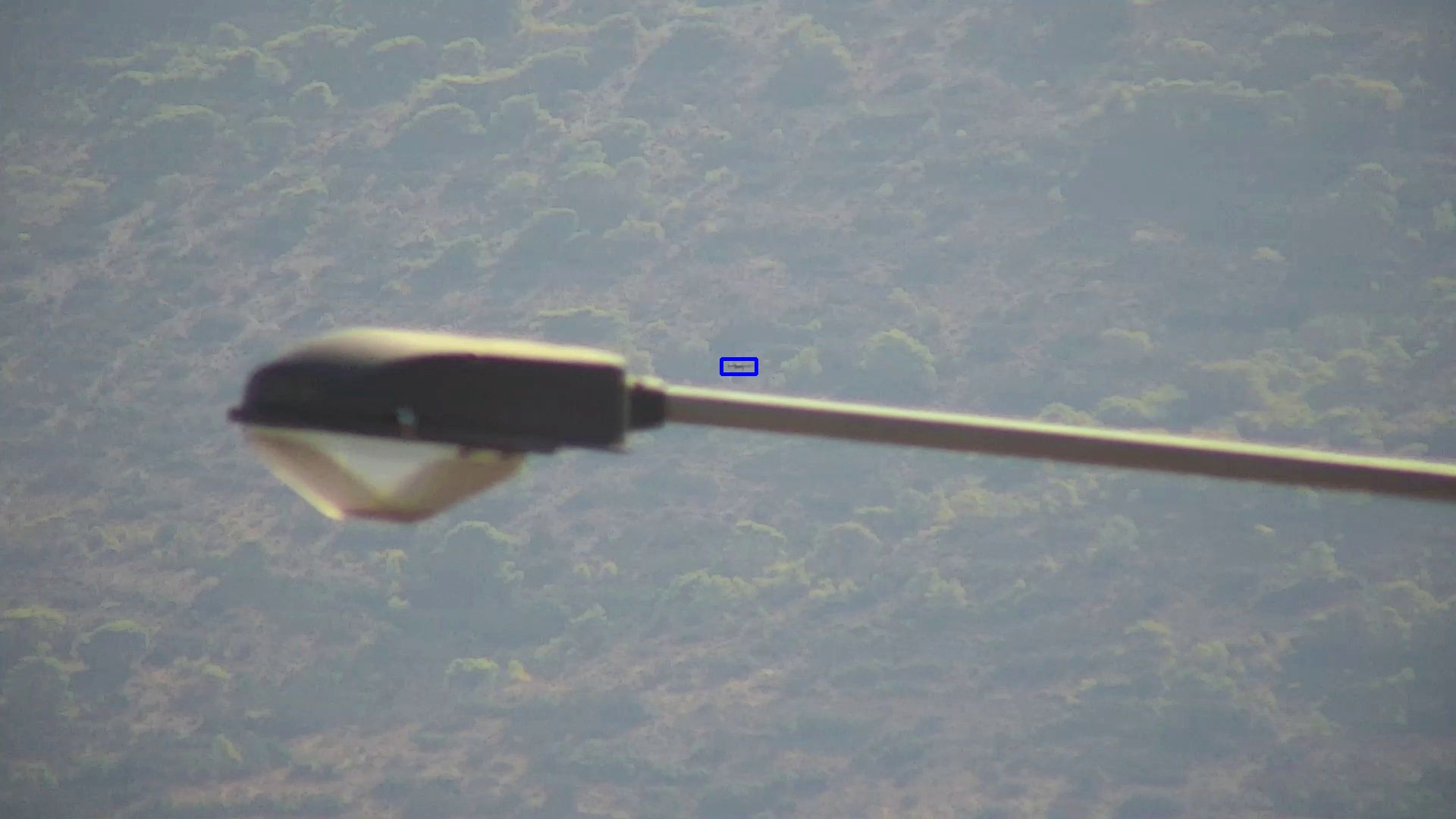}
        \includegraphics[height=14ex]{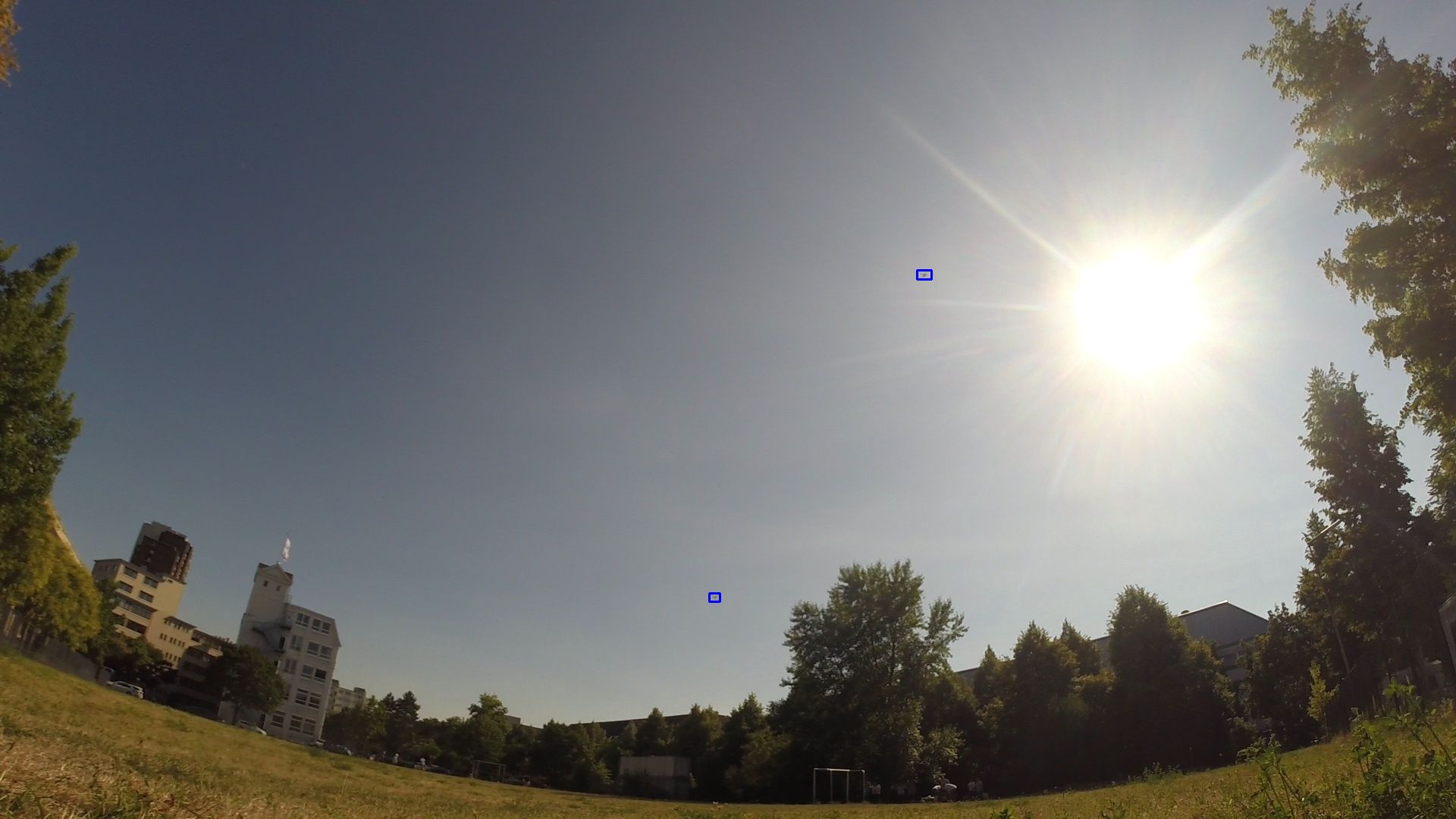}
    }

    \vspace{1mm}

    \resizebox{0.99\linewidth}{!}{
        \includegraphics[height=16.4ex]{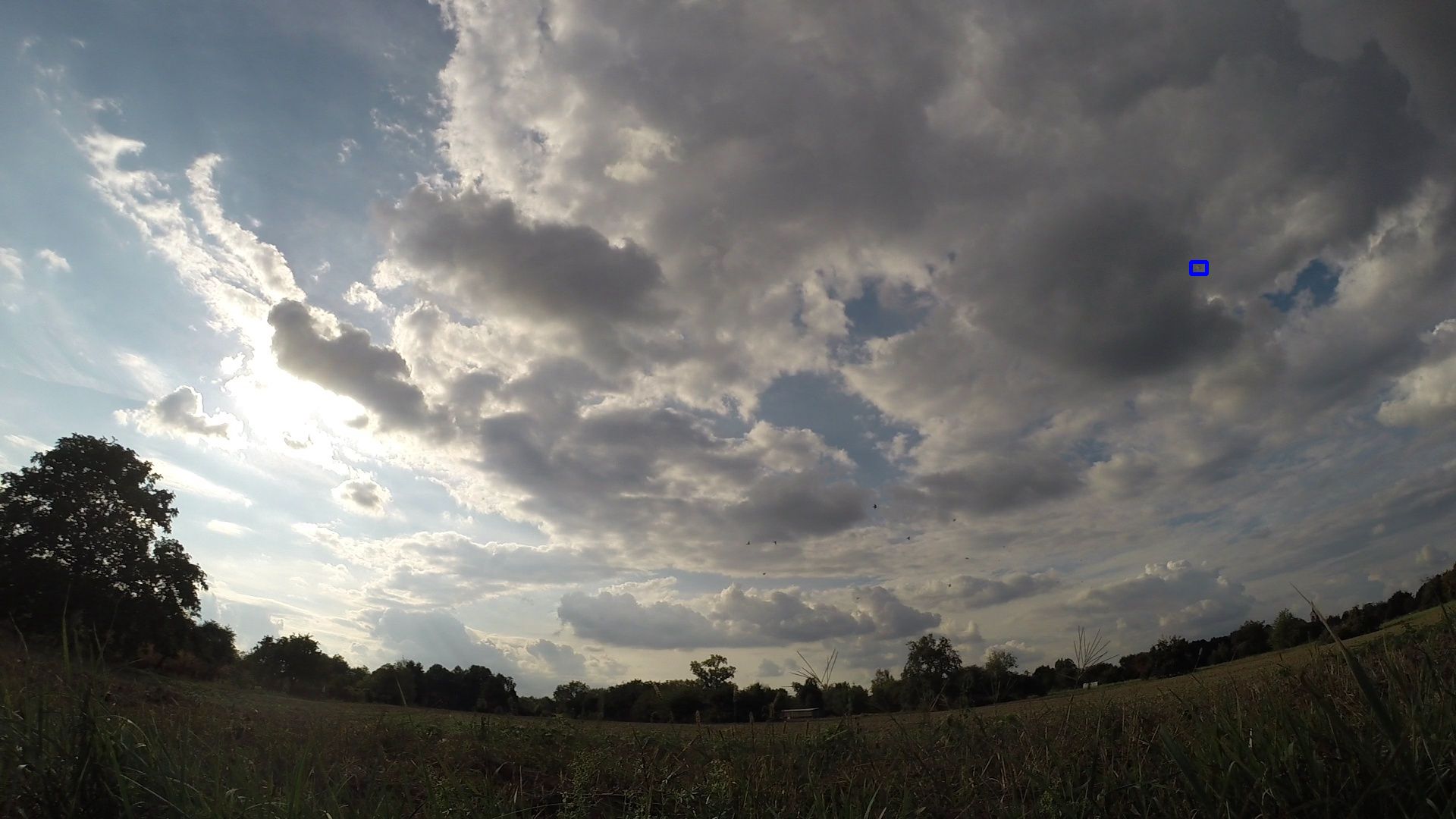}
        \includegraphics[height=16.4ex]{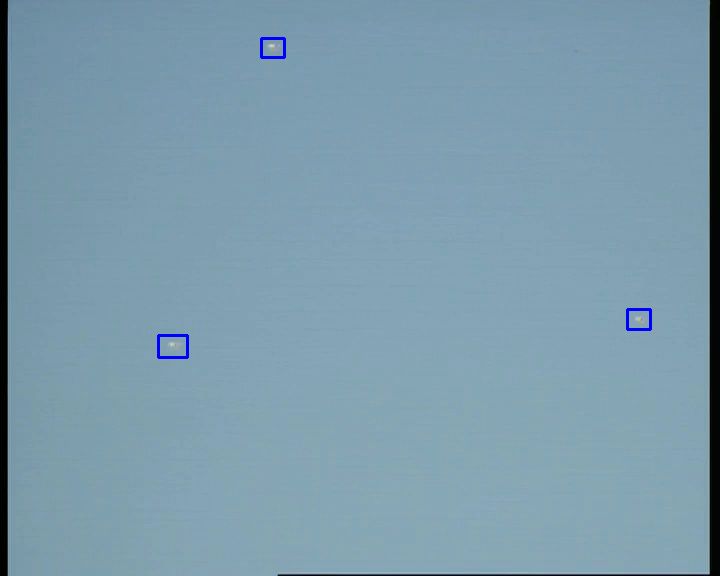}
    }

    \vspace{-2mm}
    
    \caption{Examples of drones successfully detected by the proposed approach. In this figure, we have intentionally refrained from providing zoomed-in views of distant, small drones to highlight the difficulty of detecting~them.}
    \label{fig:qualitative-results}
\end{figure}

Examples of successfully detected drones are shown in \cref{fig:qualitative-results}.
It is important to observe that the proposed approach is capable of detecting drones across various backgrounds, weather conditions, scales, among other factors.
Of particular interest is the left image in the third row of \cref{fig:qualitative-results} (also featured on the first page of this work), where none of the birds were erroneously identified as~drones.

\section{Conclusions}
\label{sec:conclusions}

In this article, we describe our approach for drone detection, which achieved first place in the \challenge~\cite{challenge}.

We utilized the medium-sized YOLOv11 model~\cite{yolov11} for drone detection.
To overcome the model's struggles with small-scale drone detection, we implemented a multi-scale approach in which the input image is processed both as a whole and in segmented components.
This strategy significantly boosted detection performance, especially for distant~drones.

To further enhance performance, we employed extensive data augmentation.
In addition to the standard transformations applied during training, we utilized a copy-paste technique to increase the number of drone and bird instances in the training images.
This involved randomly placing cropped and scaled instances into new locations, ensuring they did not overlap with existing instances.
Finally, a post-processing stage was incorporated to mitigate missed~detections.

A key performance bottleneck in our current approach is the computational overhead associated with the detector's repeated execution.
To mitigate this, we propose three primary avenues for improvement: optimizing the detection pipeline to reduce detector calls, examining faster model alternatives, and investigating the use of parallel processing or hardware acceleration to handle multiple detections~concurrently.

We aim to improve detection performance through the following investigations:
(i)~leveraging a state-of-the-art tracker to enhance multi-frame predictions;
(ii)~developing a classifier that analyzes multi-frame image patches to accurately differentiate drones from similar objects, such as birds, thereby reducing false positives~\cite{akyon2022sequence};
and (iii)~investigating latest architectures, such as D-FINE~\cite{peng2024dfine} and RT-DETRv2~\cite{lv2024rtdetrv2}, while incorporating a wider variety of datasets, including the recently introduced SynDroneVision~\cite{lenhard2025syndronevision}, into the training~process.

\balance
\section*{\uppercase{Acknowledgments}}

\iffinal
    This study was supported in part by the \textit{Coordenação de Aperfeiçoamento de Pessoal de Nível Superior}~(CAPES) -- Finance Code 001, and by the \textit{Conselho Nacional de Desenvolvimento Científico e Tecnológico}~(CNPq) under grants \#~$315409$/$2023$-$1$ and \#~$312565$/$2023$-$2$.
    We gratefully acknowledge the \textit{Pontifícia Universidade Católica do Paraná} and \textit{Fundação Araucária} for their financial support enabling conference participation.
    We also thank NVIDIA Corporation for donating the Quadro RTX $8000$ GPU used in this~research.

\else
    The acknowledgments are hidden for review.
\fi

\bibliographystyle{IEEEtran}
\bibliography{bibtex}
\end{document}